\documentclass[journal]{IEEEtran}
\ifCLASSOPTIONcompsoc
  \usepackage[nocompress]{cite}
\else
  \usepackage{cite}
\fi
\ifCLASSINFOpdf
\else
\fi
\usepackage{times}
\usepackage{helvet}
\usepackage{courier}
\usepackage{amssymb}
\usepackage{amsmath}
\usepackage{graphicx}
\usepackage{subfigure}
\usepackage{amsthm}
\usepackage{multirow}
\usepackage{booktabs}
\usepackage{epstopdf}
\usepackage{color}

\begin{document}
%
\title{Deep Recurrent Regression for Facial Landmark Detection}
%
%
%
%

\author{Hanjiang Lai, Shentao Xiao, Yan Pan, Zhen Cui, Jiashi Feng, Chunyan Xu, Jian Yin, Shuicheng Yan,~\IEEEmembership{Senior Member,~IEEE}

\IEEEcompsocitemizethanks{\IEEEcompsocthanksitem Hanjiang Lai, Yan Pan and Jian Yin are with School of Data and Computer Science, Sun Yat-Sen University, China. This work was mainly performed when Hanjiang Lai was staying in NUS.
\IEEEcompsocthanksitem Shengtao Xiao, Zhen Cui， Jiashi Feng and Shuicheng Yan are with Department of Electronic and Computer Engineering, National University of Singapore, Singapore.
\IEEEcompsocthanksitem Chunyan Xu is with School of Computer Science and Engineering, Nanjing University of Science and Technology, China.
}
\thanks{Hanjiang Lai and Shengtao Xiao contributed equally to this work. Yan Pan is the Corresponding Author (panyan@mail.sysu.edu.cn). }}

%
%

\markboth{}%
{Shell \MakeLowercase{\textit{et al.}}: Bare Demo of IEEEtran.cls for Computer Society Journals}
%



\IEEEtitleabstractindextext{%
\begin{abstract}
We propose a novel end-to-end deep architecture for face landmark detection, based on a deep convolutional and deconvolutional network followed by carefully designed recurrent network structures. The pipeline of this architecture consists of three parts. Through the first part, we encode an input face image to resolution-preserved deconvolutional feature maps via a deep network with stacked convolutional and deconvolutional layers. Then, in the second part, we estimate the initial coordinates of the facial key points by an additional convolutional layer on top of these deconvolutional feature maps. In the last part, by using the deconvolutional feature maps and the initial facial key points as input, we refine the coordinates of the facial key points by a recurrent network that consists of multiple Long-Short Term Memory (LSTM) components. Extensive evaluations on several benchmark datasets show that the proposed deep architecture has superior performance against the state-of-the-art methods.
\end{abstract}

\begin{IEEEkeywords}
Cascaded Regression, Facial Landmark Detection, Shape-indexed Features, Recurrent Regression.
\end{IEEEkeywords}}

\maketitle

\IEEEdisplaynontitleabstractindextext

%
\IEEEpeerreviewmaketitle

\section{Introduction}\label{sec:introduction}
Facial landmark detection is a task of automatically locating pre-defined facial key points on a human face. It is one of the core techniques for solving various facial analysis problems, e.g., face recognition~\cite{zhao2003face}, face morphing~\cite{liu2014wow,kemelmacher2014illumination}, 3D face modeling~\cite{cao2014displaced} and face beautification~\cite{liu2014wow}. In recent years, considerable research effort has been devoted to developing models for accurately localizing the landmark points on the face images captured under unconstrained conditions based on the provided face detection bounding box~\cite{saragih2011deformable,zhu2012face,martins2013generative}. Among these methods, a notable research line is the cascaded regression methods~\cite{cao2014face,sdm,dollar2010cascaded,lee2015face}, which have shown a strong ability to efficiently and accurately localize the facial key points even in challenging scenarios.

\begin{figure*}
\centering
    \includegraphics[width=1\hsize \hspace{0.01\hsize}]{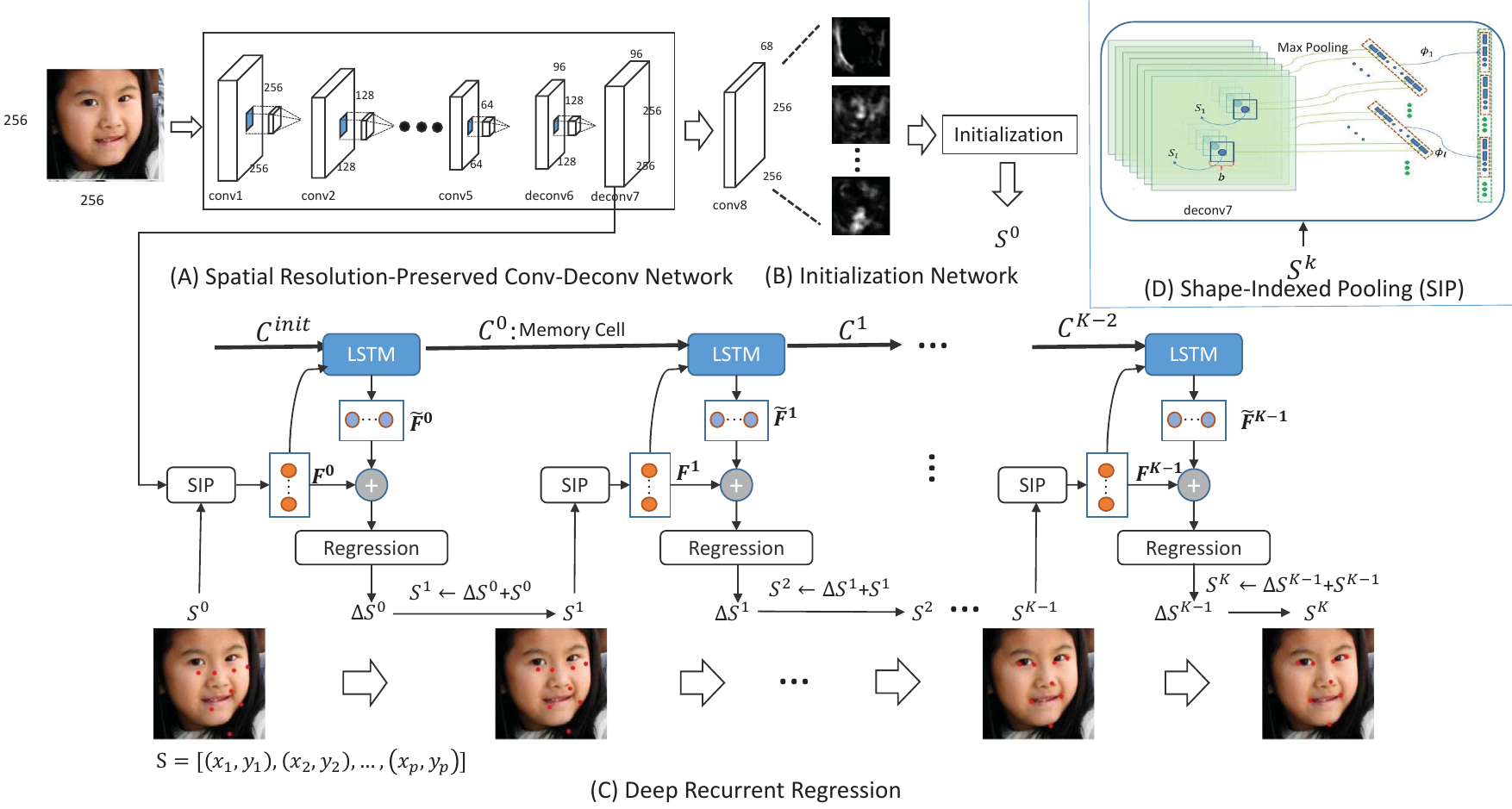}
  \caption{An overview of the proposed end-to-end deep architecture for facial landmark detection. Given an input image, our architecture (A) encodes the image into the resolution-preserved deconvolutional feature maps, e.g., \textit{deconv7}, in the same size of the image via stacked convolutional and deconvolutional layers; With this, it can highly increase the power of the feature representations. And then (B) constructs a small sub-network to provide a more reliable shape initialization for further refinement of landmarks. A convolutional layer, e.g., \textit{conv8}, is added, which has $p$ channels, each channel for predicting one facial key point. With this, roughly estimated locations of landmarks, e.g., $S^0$, can be obtained from \textit{conv8}. (C) is a deep recurrent network to refine the shape iteratively. It firstly extracts the deep shape-indexed features $F^k$ based on the \textit{deconv7} layer and the current estimated shape, via shape-indexed pooling (SIP) layer as shown in (D). Then, it learns the recurrent shape features $\widetilde{F}^{k}$ by the LSTM component. After that, with the deep shape-indexed features and the recurrent shape features as input, a sequence of regressions are conducted to iteratively refine the facial shape and obtain the final shape. To make the proposed network to be easily understood, we draw the recurrent network in the unrolled form.}
  \label{framework}
\end{figure*}

The cascaded regression methods are characterized by such a pipeline: at each cascading stage, visual features are first extracted from the current predicted landmark points; the coordinates of the landmark points are then updated via regression from the extracted features; and the updated landmark points are used for regression in the next stage. These operations are repeated for several times to iteratively refine the predicted locations of landmark points.

Despite of their acknowledged success, the performance of such cascaded regression methods heavily depends on the quality of the adopted visual features. Most of the existing cascaded regression methods employ popular hand-crafted features such as SIFT, HOG~\cite{sdm,cfss} or binary features extracted by random forest models~\cite{lbf}. However, such hand-crafted features may not be optimally compatible with the process of cascaded regression, making the learned models usually be sensitive to large occlusion, extreme poses of human faces, large facial expressions, or varying illumination conditions.

In addition, in the $k$-th cascading stage of the existing cascaded regression methods (e.g., SDM~\cite{sdm}), only the current estimated coordinates of landmarks (denoted as $S^k$) are used to conduct regression, but ignoring the landmark information estimated in more previous iterations (i.e., $S^0$, $S^1$, ..., $S^{k-1}$). Since the cascaded regression iteratively refines a set of estimated landmarks, these estimated sets of landmarks (i.e., $S^0$, $S^1$, ..., $S^{k-1}$, $S^{k}$) are expected to have high correlation to each other. We argue that, in the $k$-th cascading stage, if we could effectively incorporate all of $S^0$, $S^1$, ..., $S^{k-1}$, $S^{k}$ into learning, we would improve the performance of regression.

In this paper, we propose a deep end-to-end architecture that iteratively refines the estimated coordinates of the facial landmark points. This deep architecture can be regarded as a ``mimicry'' of the cascaded regression based on deep networks. The key insight of the proposed architecture is to combine the strength of convolutional neural networks (CNNs) for learning better feature representation and recurrent neural networks (RNNs) for memorizing the previous estimated information of landmark points. Specially, as shown in Fig.~\ref{framework},  this architecture has three building blocks: (1) an input face image is firstly encoded to resolution-preserved deconvolutional feature maps, each is in the same size of the input image, via stacked convolutional and deconvolutional layers. (2) On top of these deconvolutional feature maps, we construct a small sub-network to estimate the initial coordinates of the facial landmark points by adding an additional convolutional layer. Such estimated initial facial landmarks provide a good starting point for the further refinement of the estimated landmarks. (3) With the deconvolutional feature maps and the initial facial landmarks, we construct a carefully designed recurrent network with multiple Long-Short Term Memory (LSTM) components to iteratively refine the estimated coordinates of the facial key points. This recurrent network can be regarded as a ``mimicry'' of the pipeline of cascaded regression methods. For each LSTM component, visual features are directly extracted from the deconvolutional feature maps via the proposed shape-indexed pooling (SIP) layer, and these visual features are used as input to the LSTM component. The LSTM components are connected one by one, the current one outputs a ``recurrent feature vector'' to the next one, where this ``recurrent feature vector'' can be regarded as the memory of the estimated information of landmarks in the previous LSTM components.

Our contributions in this paper can be summarized as follows:
\begin{itemize}
\item We propose a novel recurrent network that can iteratively refine the estimated coordinates of the facial key points. This recurrent network can be regarded as a ``mimicry'' of the pipeline of cascaded regression methods. This recurrent network has multiple LSTM components that are connected one by one, the current one outputs a ``recurrent feature vector'' to the next one, where this ``recurrent feature vector'' can be regarded as the memory of the estimated information of landmarks in the previous LSTM components.

\item We develop a deep convolution-deconvolution network that can encode an input face image to resolution-preserved feature maps, where these feature maps can be used to estimate accurate initial coordinates of the facial landmark points, or extract powerful visual features as input to the above mentioned recurrent network. This deep convolution-deconvolution network and the above mentioned recurrent network are connected to be an end-to-end architecture that can be trained by back propagation.

\end{itemize}

The rest of the paper is organized as follows. First, a brief review of related work will be given in Section \ref{related_work}. Then we will show an end-to-end deep architecture for facial landmark detection in Section \ref{approach}. The experiments and analysis are reported in Section \ref{experiments}. Finally, the conclusions are drawn in Section \ref{conclusion}.

\section{Related Work}
\label{related_work}
%

\subsection{Cascaded Regression Methods}
The most related works are cascaded regression methods. The comprehensive survey of other methods for facical landmark detection, e.g, CLM-based methods~\cite{martinez2013local} and AAM-based methods~\cite{gao2010review} can be found in \cite{wangsurvey}. 

Cascaded regression methods are a representative stream of facial landmark detection methods. They usually start from a set of initial coordinates of facial landmark points, and then iteratively refine the estimation of these landmark points. A representative method of the cascaded regression methods is the supervised descent method (SDM)~\cite{sdm}. SDM uses SIFT features extracted around the current landmarks and solves a series of linear least square problems to iteratively refine these landmarks. Ren \textit{et al.}~\cite{lbf} propose the locality principle to learn a set of local binary features for cascade regression. Global SDM (GSDM)~\cite{globalsdm} is an extension to SDM which divides the search space into regions of similar gradient directions.

In the last few years, we are witnessing dramatic progress in deep convolutional networks. In contrast to the methods that use hand-crafted visual features (e.g, SIFT, LBP), deep-convolutional-networks-based methods can automatically learn discriminative visual features from images. Sun \textit{et al.}~\cite{sun2013deep} propose an approach to predict facial key points with three-level convolutional networks. Liu \textit{et al.}~\cite{liu2015dual} propose a dual sparse constrained cascaded regression model for robust facial landmark detection. Zhang \textit{et al.}~\cite{zhang2014coarse} propose a Coarse-to-Fine Auto-encoder Networks (CFAN) for facial alignment. Zhang \textit{et al.}~\cite{zhang2014topic} propose a topic-aware face alignment method to divide the difficult task of estimating the target landmarks into several much easier subtasks. DRN-SSR~\cite{zhang2015leveraging} is a deep regression network coupled with sparse shape regression to predict the union of all types of landmarks by leveraging datasets with varying annotations. Belharbi \textit{et al.}\cite{belharbi2015facial} formulate the face alignment as a structured output problem and exploit the strong dependencies between the outputs.

The performance of cascaded regression methods may heavily depend on the quality of initial face landmarks in the testing stage. Several methods have been developed to  obtain a good initialization of face landmarks, including multiple random shape initializations~\cite{cao2014face}, smart restarts~\cite{burgos2013robust} and coarse-to-fine searching~\cite{cfss,sun2013deep,zhang2014coarse}. Cao \textit{et al.}~\cite{cao2014face} run the cascaded regression method several times, each with a different initial shape, and take the median results as the final landmark estimation. Zhu \textit{et al.}~\cite{cfss} propose a coarse-to-fine searching method that begins with a coarse search over a shape space with diverse shapes, and employs the coarse solution
to constrain subsequent finer search of shapes. Zhang \textit{et al.}~\cite{zhang2014coarse} propose a coarse-to-fine auto-encoder network to find the initial shapes.

Different from the conventional cascaded regression approach, in this paper, we propose a novel recurrent network that can iteratively refine the estimated facial landmarks, where this recurrent network can be regarded as a ``mimicry'' of the pipeline of cascaded regression methods.

We develop a carefully-designed deep convolution-deconvolution network that encodes an input face image to resolution-preserved deconvolutional feature maps. The visual features for facial landmark detection are directly extracted from these feature maps. This shares some similarities to the deep-convolutional-networks-based methods (e.g.,~\cite{sun2013deep,zhang2014coarse,liu2015dual}) but we use a quite different deep architecture.

Different from the existing initialization strategies for facial shapes in the testing stage, by adding a simple sub-network on top of the deconvolutional feature maps, we obtain initial face shapes from the output of this sub-network.

\subsection{Recurrent Neural Networks}

Recently, Recurrent Neural Networks (RNNs) have received a lot of attention and achieved impressive results in various applications, including speech recognition~\cite{lstm_speech}, image captioning~\cite{showtell} and video description~\cite{videoCVPR}. RNNs have also been applied to facial landmark detection. Very recently, Peng et al.~\cite{peng2016recurrent} propose a recurrent encoder-decoder network for video-based sequential facial landmark detection. They use recurrent networks to align a sequence of the same face in video frames, by exploiting recurrent learning in spatial and temporal dimensions.

Different to the method in~\cite{peng2016recurrent} for video-based sequential facial landmark detection, in this paper, we focus on facial landmark detection in still images and use recurrent networks to iteratively refine the estimated facial landmarks.

In addition, the proposed recurrent network in this paper contains multiple Long-Short Term Memory (LSTM) components. For self-containness, we give a brief introduction to LSTM. LSTM is one type of the RNNs, which is attractive because it is explicitly designed to remember information for long periods of time. LSTM takes $x_t$, $h_{t-1}$ and $c_{t-1}$ as inputs, $h_t$ and $c_t$ as outputs:

\begin{equation}
\nonumber
\begin{aligned}
 &i_t = \sigma(W_i[x_t;h_{t-1}] + b_i)& \\
 &f_t = \sigma(W_f[x_t;h_{t-1}] + b_f)& \\
 &o_t = \sigma(W_o[x_t;h_{t-1}] + b_o)& \\
 &g_t = \text{tanh}(W_g[x_t;h_{t-1}] + b_g)&\\
 &c_t = f_t \odot c_{t-1} + i_t \odot g_t&\\
 &h_t = o_t \odot c_t,&
\end{aligned}
\end{equation}
where $\sigma(x) = (1+e^{-x})^{-1}$ is the sigmoid function. The outputs of the sigmoid functions are in the range of $[0,1]$, here the smaller values indicate ``more probability to forget the previous information'' and the larger values indicate ``more probability to remember the information''. $\text{tanh}(x) = \frac{e^x - e^{-x}}{e^x + e^{-x}}$ is the tangent non-linearity function and its outputs are in the range $[-1,1]$. $[x;h]$ represents the concatenation of $x$ and $h$. LSTM has four gates, a forget gate $f_t$, an input gate $i_t$, an output gate $o_t$ and an input modulation gate $g_t$. $h_t$ and $c_t$ are the outputs. $h_t$ is the hidden unit. $c_t$ is the memory cell that can be learned to control whether the previous information should be forgotten or remembered.

\section{The Approach}
\label{approach}

The objective function of the traditional cascaded regression methods~\cite{sdm} can be formulated as
\begin{equation}
 \min_{\Delta S^k} ||\phi(S^k + \Delta S^k) - \phi(S^*) ||_2^2, k = 0,\cdots,K-1,
\label{cascade_formulation}
\end{equation}
where $S$ is in the form of $S = \{(x_1,y_1),(x_2,y_2),...,(x_p,y_p)\}$ that $(x_i,y_i)$ represents the coordinate of the $i$-th facial key point ($i=1,2,...,p$). $S^{*}$ is the ground truth and $S^0$ is the initial configuration of the landmarks (e.g., the mean shape of all training images). We denote $\phi(S)$ as the \textit{shape-indexed feature} extracted from the shape $S$. $\Delta S$ is the shape increment, which is needed to be learned. The goal of cascaded regression is to generate a sequence of updates ($\Delta S^{0}, \ldots, \Delta S^{K-1}$) starting from $S^0$ and converges to $S^{*}$ (i.e., $S_0 + \sum_{k=0}^{K-1} \Delta S^k \approx S^{*}$).

For the traditional cascaded regression method, only the current shape $S^k$ is considered in the $k$-th regression, little information about the previous shapes (e.g., $S^0,S^1,\cdots,S^{k-1}$) is kept. We argue that the cascade regression should be able to connect previous shape information to the present shape regression.  We introduce a new feature $\widetilde{F}^{k}$~\footnote{Note that $\widetilde{F}^{k}$ represents the recurrent shape features that are obtained from the the previous and current shape information (i.e.,$F^0,F^1,\cdots,F^k$), hence we use the index $k$ (instead of $k-1$) for $\widetilde{F}$.} into (\ref{cascade_formulation}). The objective function is changed to
\begin{equation}
\min_{\Delta S^k}||\phi(S^k + \Delta S^k) + \widetilde{F}^{k} - \phi(S^*) ||_2^2.
\end{equation}

The new feature is referred as \textit{recurrent shape feature}, which is used to incorporate the previous information and help to improve the current prediction. Due to LSTMs have the strong ability of learning long-term dependencies, we utilize the LSTM to learn the  recurrent shape feature (please refer SubSection \ref{drrn} for more details).

Let $F^k = \phi(S^k)$ and $f(S^k + \Delta S^k) = ||\phi(S^k + \Delta S^k) + \widetilde{F}^{k} - \phi(S^*) ||_2^2$. Similar to SDM [9], we also assume that $\phi$ is twice differentiable under a small shape change. The Taylor expansion of $f(S^k + \Delta S^k)$ is
\begin{equation}
f(S^k + \Delta S^k) \approx f(S^k) +  f^{'}(S^k)^T \Delta S^k + \frac{1}{2} (\Delta S^k)^T  f^{''}(S^k) \Delta S^k,
\end{equation}
where $f^{''}(S^k)$ and $f^{'}(S^k)$ are the Hessian and Jacobian matrices of $f$ at $S^k$. Now the optimization objective becomes
\begin{equation}
\min_{\Delta S^k} f(S^k) +  f^{'}(S^k)^T \Delta S^k + \frac{1}{2} (\Delta S^k)^T  f^{''}(S^k) \Delta S^k,
\label{loss_delta}
\end{equation}
where $\Delta S^k$ is the variable. By letting the derivatives of the objective in (\ref{loss_delta}) with respect to $\Delta S^k$ to be zeros, we have
\begin{equation}
\Delta S^k = - [f^{''}(S^k)]^{-1} f^{'}(S^k).
\label{update_delta}
\end{equation}

According to the chain rule, we have $f^{'}(S^k) = \frac{\partial f}{\partial S^k} = \frac{\partial f}{\partial \phi} \times \frac{\partial \phi}{\partial S^k} = 2 \times (\phi(S^k) + \widetilde{F}^{k} - \phi(S^{*})) \times \phi^{'}(S^k) $. Substituting it into (\ref{update_delta}), we have
\begin{equation}
\begin{aligned}
\Delta S^k = & - [f^{''}(S^k)]^{-1} f^{'}(S^k) & \\
= & - [f^{''}(S^k)]^{-1} \times 2\phi^{'}(S^k)  \times (F^k + \widetilde{F}^{k} - \phi(S^{*})).&
\end{aligned}
\end{equation}

Let $R^k = -2[f^{''}(S^k)]^{-1} \times \phi^{'}(S^k) $ and $b^k = 2[f^{''}(S^k)]^{-1} \phi^{'}(S^k) \phi(S^{*})$, then $\Delta S^k$ can be written as
\begin{equation}
\Delta S^k = R^k(F^k + \widetilde{F}^{k}) + b^k,
\label{update_rule}
\end{equation}
which is the linear combination of the current shape-indexed feature $F^k$ and the recurrent shape feature $\widetilde{F}^{k}$ plus a biased term $b^k$. Now the objective function can be formulated as
 \begin{equation}
 \begin{aligned}
 & \min_{R^k,b^k} ||S^* - S^k - R^k(F^k +\widetilde{F}^{k}) - b^k ||^2,& \\
 & \ \ \ k=0,\cdots,K-1.&
  \end{aligned}
  \label{objective}
 \end{equation}

Four key issues should be addressed in our objective (\ref{objective}):
\begin{itemize}
\item How to find the good initial shape $S^0$?
\item How to design strong deep shape-indexed feature $F^k$?
\item How to keep a state or memory along the sequence by the recurrent shape feature $\widetilde{F}^{k}$?
\item How to update the linear parameters $R^k$ and bias $b^k$?
\end{itemize}

In this paper, we propose a deep architecture designed for facial landmark detection as shown in Fig.~\ref{framework}. Given an input face image, the pipeline of the proposed architecture contains three parts: 1) a spatial resolution-preserved conv-deconv network with multiple convolutional-pooling layers followed by two deconvolutional layers, which is to obtain powerful feature maps; 2) a small initialization network to predict the initial coordinates of the facial key points by adding an additional convolutional layer on top of the deconvolutional feature maps; This module is to find good initial shape $S^0$. 3) a recurrent network with LSTM component to refine the coordinates of the facial key points, which includes the SIP layer for extracting deep shape-indexed feature $F^k$, the LSTM component for learning the recurrent shape feature $\widetilde{F}^{k}$ and sequential linear regressions for learning a serial of $R^k$ and $b^k$, $k=0,\cdots,K-1$. In the following, we will present the details of these parts, respectively.

\subsection{Spatial Resolution-Preserved Conv-DeConv Network.} 
\begin{figure}
    \includegraphics[width=1\hsize \hspace{0.01\hsize}]{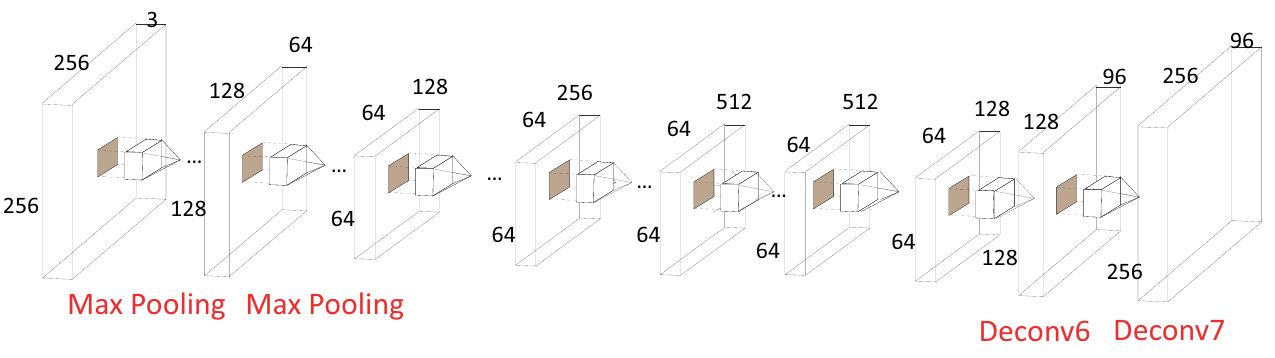}
  \caption{Illustration of deep spatial resolution-preserved conv-deconv network based on the variant of the VGG-19. Max pooling layers are used to down-sampling and deconvolutional layers are used to up-sampling.}
  \label{subnetwork}  
\end{figure}

We propose a spatial resolution-preserved conv-deconv network to learn the powerful feature maps that facilitate the following shape feature extraction and shape regressions. Our network is built on the VGG-19 layer net~\cite{vgg} as shown in Fig.~\ref{subnetwork}, where makes following structural modifications. The first modification is to remove the last three max pooling layers (pool3, pool4, pool5) and all the fully-connected layers (fc6, fc7, fc8). The second is to add two deconvolutional layers \cite{long2014fully,noh2015learning} (deconv6, deconv7).  \textit{Deconvolution}, which is also called \textit{backwards convolution}, reverses the forward and backward passes of the convolution. It is used for up-sampling. Our conv-deconv network has 16 convolutional layers, 2 max pooling layers and 2 deconvolutional layers. In the convolutional layers, we zero-pad the input with $\lfloor k/2  \rfloor$  zeros on all sides, where $k$ is the kernel size of filters in a specific layer. By this padding, the input and the output can have the same size. The pooling layer filters the input with a kernel size of $2 \times 2$ and a stride of 2 pixels, which makes the size of the output be half of the input. In the deconvolutional layer, the input is filtered with 96 kernels of size $4 \times 4$ and a stride size of 2 pixels, which makes the size of the output be 2 times of the input. Suppose that the input image size is $H \times W$. After passing through two pooling layers, the image becomes $H/4 \times W/4$, and then is upsampled to $H \times W$ via two deconvolutional layers.

In such modifications, the raw image is changed to the feature maps of the same size of the image, which can obtain more powerful representation and help to get more accurate results.

\textbf{Discussions.} The proposed network architecture is based on VGG-19, which is a widely used network for various vision tasks (e.g., object classification). Hereafter we assume that the fully connected layers in VGG-19 are removed. VGG-19 reduces the size of activations by repeatedly using max pooling operations. Too many pooling operations may make the output feature maps too small to keep detailed spatial information. However, in the task of facial keypoint detection, we need output feature maps that contain sufficiently detailed and accurate spatial information. For instance, for an $224 \times 224$ image as input to VGG-19, after going through $4$ max pooling layers, the size of each of the output feature maps (conv$5\_4$) is $14 \times 14$. Suppose two facial key points have a distance less than $10$ pixel in the input image (e.g., two points in the left eye), then these facial points may highly probably be mapped to the same ``pixel'' in the $14\times 14$ output feature maps, making it very difficult to discriminate these points.

To address this issue, a straightforward alternative is an architecture that removes all of the pooling layers in VGG-19 and keeps other layers uncharged. In such a variant of VGG-19, the size of the output feature maps (conv$5\_4$) is the same as the input image. It can keep detailed spatial information in the output feature maps. However, it can be verified that, compared to VGG-19, such a variant (without pooling layers) has considerably higher time complexity in training or conducting predictions because the output feature maps in the intermediate layers are larger than those of VGG-19.

The main goals behind the design of the proposed deep architecture (VGG-S) are two-fold: (1) The size of the output feature maps is sufficiently large so as to keep detailed spatial information. (2) The time complexity of training/testing is acceptable. Specifically, in the proposed architecture, we remove the last two (of four) pooling layers~\footnote{Note that the fifth max pooling layer (pool5) is after the last convolutional layer (conv$5\_4$). } in VGG-19, and then add two deconvolutional layers after conv$5\_4$. We use the features maps on the second deconvolutional layer (deconv7) as the output feature maps for facial keypoint detection, where the size of the feature maps in deconv is the size as the input image. These feature maps provide sufficiently detailed spatial information for facial keypoint detection. Simultaneously, the time complexity of the proposed architecture in training/testing is only slightly higher than that of VGG-19 (without fully connected layers).

\subsection{Initialization Network}
The good initialization for cascaded regression methods is very important, which have been indicated by~\cite{cfss,smith2014nonparametric}. Many algorithms, e.g.~\cite{sdm}, use the mean shape as the initialization. In this paper, we propose a simple method to find the initialization from the shape space instead of using a specified initialization.

After the image goes through the spatial resolution-preserved network, it is mapped to high-level features maps (deconv7).  We add a new layer (i.e., conv8) which has $p$ channels, each channel for predicting one facial key point. We refer them as the local mapping functions. The local mapping functions can give us a predicted shape. Also we construct $N$ candidate shapes, which cover different poses, expression and so on. Given the predicted shape, the good initialization from the set of candidate shapes should have the minimum distance between the predicted shape.

Note that we do not directly use the predicted shape from the local mapping functions as the initial shape. The reason is that the local mapping functions only consider the local information (e.g, each function only consider one key point), thus sometimes the predicted shape may not look like a face.

\begin{figure} [t]
\centering
    \includegraphics[width=1\hsize \hspace{0.01\hsize}]{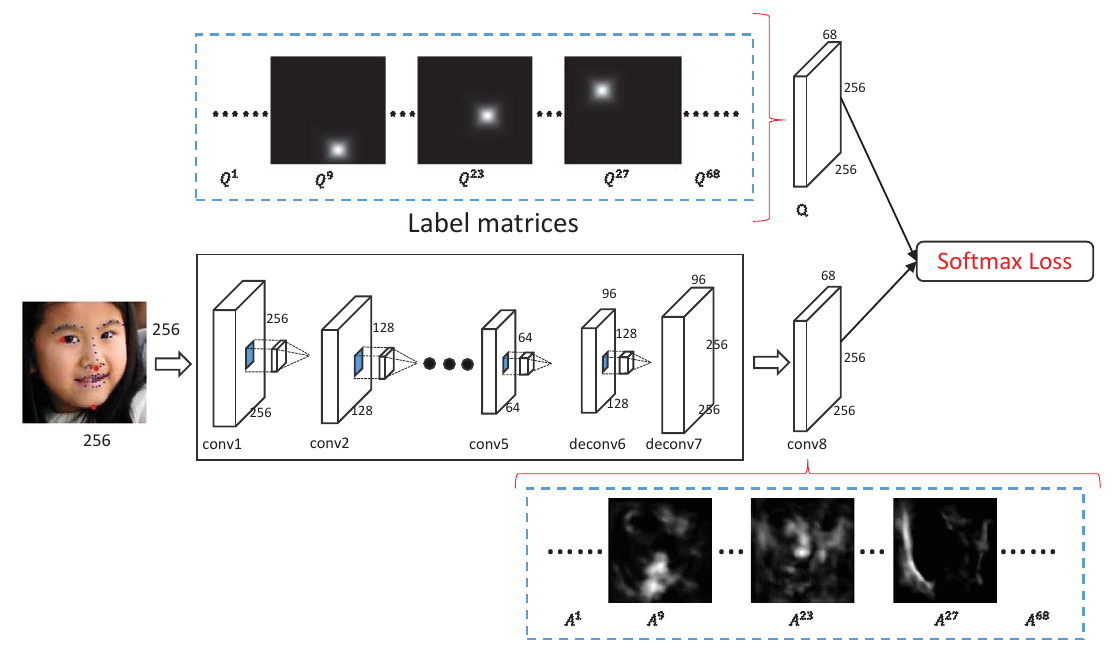}
  \caption{Learning local mapping functions. The middle part shows our initialization network for the input image with 68 landmarks, e.g., the red points in the facial images are the $9^{th}$, $34^{th}$ and $37^{th}$ landmarks, respectively. After the input image goes through the spatial resolution-preserved initialization network, the network will output a conv8 layer that has 68 features maps, each corresponds to one landmark. The top part shows the 68 ground truth probability matrices. Each grey image represents a ground truth probability matrix that corresponds to one landmark, where whiter color indicates higher possibility of the landmark existing at that position. In training, softmax loss is used to measure the probability-distance between the $i$-th feature map in conv8 and the $i$-th ground truth $Q^i$ in the top part ($i=1,2,...,68$). }
  \label{local_mapping}  
\end{figure}

\textbf{Learning local mapping functions.}
Now the problem is how to find the predicted shape. Since the size of conv8 is the same as  the input image, a pixel indexed $(x,y)$ in the conv8 can be mapped to the same pixel in the input image. We make the $i$-th feature map predict the location of the $i$-th facial key point as follow: the location of the largest value in the $i$-th feature map is the $i$-th facial key point.

We denote the $i$-th feature map of conv8 as $A^i \in \mathbb{R}^{H \times W}$ and the $i$-th facial key point as $(x_i,y_i)$, where $H$ / $W$ is the height / width of the feature map. Let $\hat{A}^i_{jk} =  \frac{e^{A^i_{jk}}}{Z}$, where $Z=\sum_{jk} e^{A^i_{jk}}$. $\hat{A}^i$ is the probability matrix, and $\hat{A}^i_{jk}$ indicates the probability of this pixel  belonging to the landmark.

A good possibility matrix should preserve the following information: (1) the probability for the index $(x_i,y_i)$ should be the largest; and (2) the farther it is away from $(x_i,y_i)$, the smaller the probability should be. Therefore, we introduce a new ground truth probability matrix $Q^i \in \mathbb{R}^{H \times W}$, which is calculated as $Q^i_{jk} = 0.5^{\max(|x_i - j|,|y_i - k|)}$ and satisfies the above two principles. Finally, a normalization process is calculated as to ensure $\sum_{jk} Q^i_{jk} = 1$.

Since $Q^i$ and $\hat{A}^i$ are two probability matrices, we propose to use the softmax loss to quantify the dissimilarity between the predicted probability matrix $\hat{A}^i$ and the ground truth probability matrix $Q^i$, which is defined as
\begin{equation}
 \min - \sum_{jk} Q^i_{jk} \log( \hat{A}^i_{jk}).
\end{equation}

Note that the local mapping functions can not only help find the initialization, but also help learn the parameters of our deep network. Since the predicted shape is obtained by only using the local information, it needs to be refined for the more accurate results.

\begin{figure}[h]
    \includegraphics[width=0.9\hsize \hspace{0.01\hsize}]{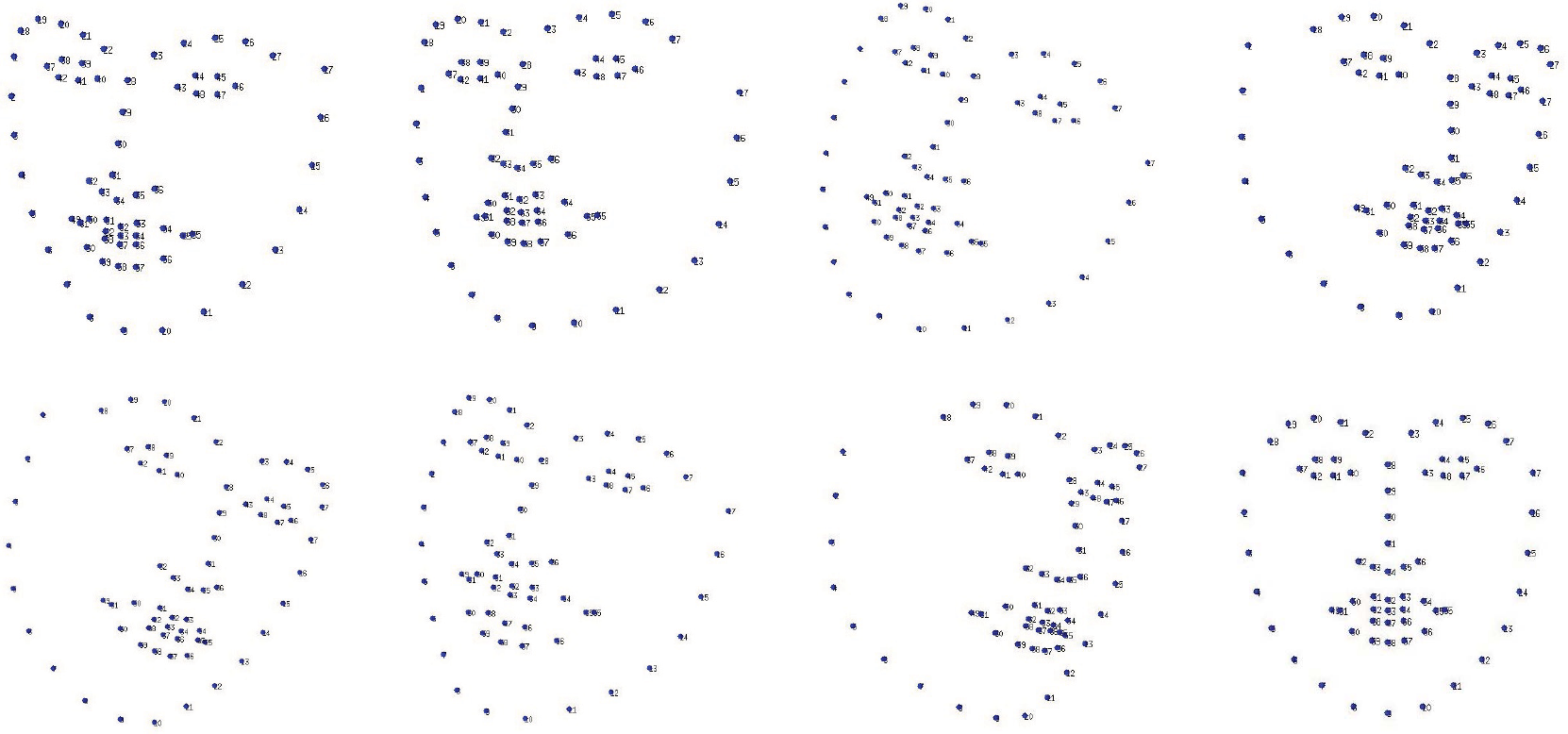}
  \caption{Several example candidate shapes from shape space.}
  \label{shapes}  
\end{figure}

\textbf{Initialization Searching.}
We firstly construct $N$ candidate shapes $\{S_1, S_2, \cdots, S_N \}$, which cover a wide range of the shapes including different poses, expressions, etc. To obtain these candidate shapes, we simply run k-means on the training set to find $N$ representative shapes. Fig. \ref{shapes} shows some example shapes.

A good initialization should be close to the ground truth shape, whereas in the testing image the latter is unknown. In this paper, we use the local mapping functions to give the predicted shape.  Suppose the predicted shape is $S$. Then we can find the initialization as
\begin{equation}
S^0 = \arg \min_{i=1,\cdots,N} ||S_i - S||^2.
\end{equation}

\subsection{Deep Recurrent Regression Network} \label{drrn}
In this module, we have the deconvolutional feature maps (deconv7) and the initial facial key points ($S^0$). We first show how to generate the shape-indexed feature $F^k$ and the recurrent shape feature $\widetilde{F}^{k}$, then show how to update the parameters $R^k$ and $b^k$.

\textbf{Extracting the Deep Shape-indexed Features.} Traditional shape-indexed features need two inputs: the original image and the current estimated shape. With the learned feature maps, we need a new way to extract the deep features. In this paper, we propose to extract the features similarly to \cite{spp}, which is referred to as ``Shape-Indexed Pooling" layer (SIP).

\begin{figure}[t]
\center
    \includegraphics[width=0.8\hsize \hspace{0.01\hsize}]{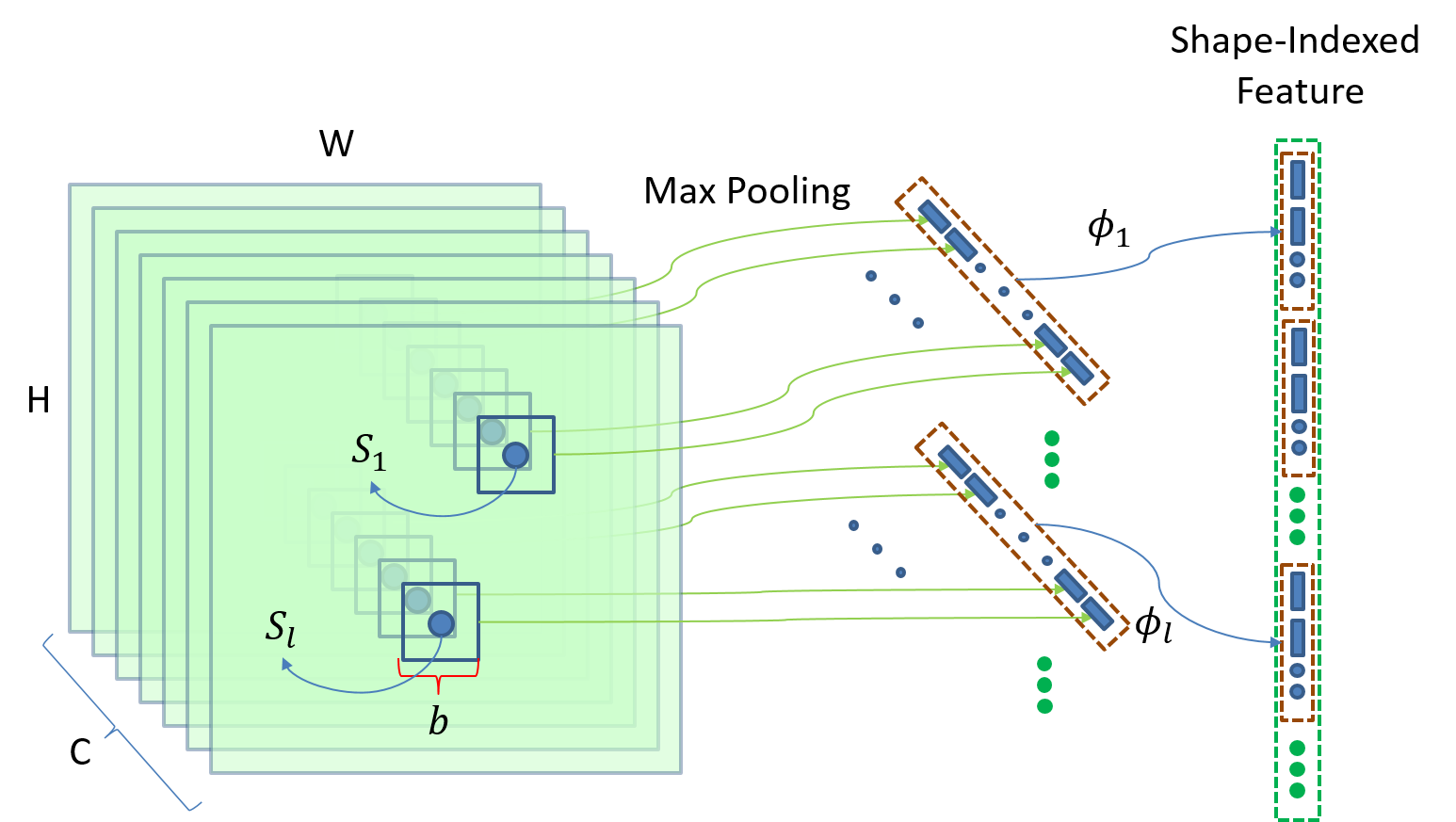}
  \caption{Shape-indexed pooling layer. We extract the deep shape-indexed features based on the current estimated shape S and the deconv7 layer. At each landmark location $(x_i,y_i) \in S$ and
the $c$-th channel, SIP encodes a region centered at $(x_i,y_i)$ with width $b$. Max-pooling is then performed within the selected region to generate the pooling response of this channel. Similar procedures are conducted for all $C$ channels and $p$ landmark points. In the end, a shape-indexed feature vector $F \in \mathbb{R}^{Cp}$ is generated by concatenating all $p$ individual landmark features.}
  \label{deep_features}  
\end{figure}

The SIP layer requires two inputs. The first input is feature maps. Note that different from SIFT and HOG which take an image as input, our proposed method uses the deep network to encode the image into high level descriptors, which gives more powerful representation. In this paper, we choose the deconv7 as the feature layer. The second input to the SIP layer is the current estimated shape. We extract the local features for all landmarks. Fig.~\ref{deep_features} shows the network structure for extracting the features based on the shape $S^k$. More specially, for each point $(x_i,y_i) \in S^k$, we localize it through a bounding box where its top-left and bottom-right coordinates are  $[x_i-b/2,y_i-b/2, x_i+b/2, y_i+b/2 ]$, with $b$ representing width of the box. We pool the responses of each filter in the local region  $[ x_i-b/2,y_i-b/2, x_i+b/2, y_i+b/2 ]$ using the max pooling. Hence, the output for each point is $C$-dimensional vector and $C$ is the number of filters in the deconv7 layer. For all the $p$ landmarks, we concatenate all the vectors into a long $Cp$-dimensional vector.

SIP is attractive because it allows end-to-end training and the features are learnable. It is different from SIFT or HOG features that are always fixed. Now, we will derive the backward propagation for the SIP layer. Let us firstly initialize the gradients to be zeros. In forward propagation, the local region $[x_i-b/2,y_i-b/2, x_i+b/2, y_i+b/2]$ is reduced to a single value. The single value has a gradient from the next layer. Then, we can just put the gradient to the place where the single value came from. Formally, let the indexed of maximum value is $(x^*,y^*)$ and the gradient of the single value in next layer is $g$, then the gradient at $(x^*,y^*)$ is added by $g$. 

\textbf{Recurrent Shape Features.} \label{RSF}
Recently, RNNs have attracted significant attention in modelling sequential data, which the networks have loops and allow information to persist. Due to LSTMs have the strong ability of learning long-term dependencies, we utilize the LSTM to learn the  recurrent shape feature. LSTMs take $F^k$, $\widetilde{F}^{k-1}$ and $C^{k-1}$ as inputs, and output $\widetilde{F}^{k}$ and $C^{k}$. The LSTM updates the inputs as following:
\begin{equation}
\begin{aligned}
 &i^k = \sigma(W_i[F^k;\widetilde{F}^{k-1}] + b_i)& \\
 &f^k = \sigma(W_f[F^k;\widetilde{F}^{k-1}] + b_f)& \\
 &o^k = \sigma(W_o[F^k;\widetilde{F}^{k-1}] + b_o)& \\
 &g^k = \text{tanh}(W_g[F^k;\widetilde{F}^{k-1}] + b_g)&\\
 &C^{k} = f^k \odot C^{k-1} + i_t \odot g^k&\\
 &\widetilde{F}^{k} = o^k \odot C^{k}.&
\end{aligned}
\label{mc}
\end{equation}
$\widetilde{F}^{k}$ is recurrent shape feature which can remember the all $k$ shapes' information.

\begin{figure}[h]
\centering
    \includegraphics[width=1.0\hsize \hspace{0.01\hsize}]{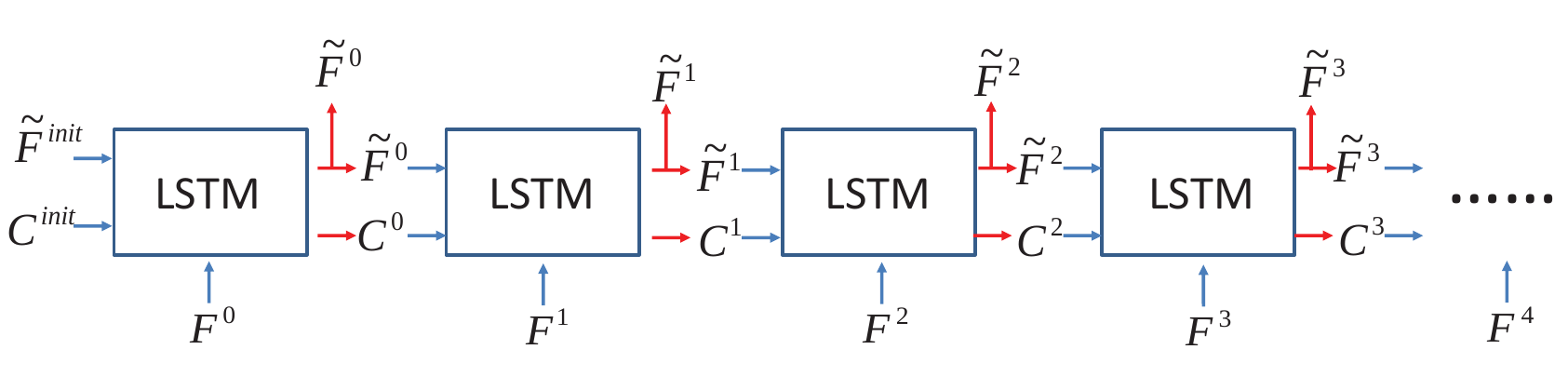}
  \caption{The unrolled version of LSTM. Note that, for each LSTM block, the $\widetilde{F}^{k}$ in the top is the same vector as the $\widetilde{F}^{k}$ in the right.  }
  \label{lstm2}  
\end{figure}

Fig. \ref{lstm2} shows the proposed LSTM network for facial landmark detection. For easily understood, we show the LSTM in unrolled form, which it is rewritten as multiple copies of the same LSTM network and each LSTM network shares the same parameters. Note that we use feed-forward connections to replace all recurrent connections. More specially, the unrolling procedure is:
\begin{equation}
\begin{aligned}
&[\widetilde{F}^0,C^0] = LSTM(F^0, \widetilde{F}^{init},C^{init})& \\
&[\widetilde{F}^1,C^1] = LSTM(F^1, \widetilde{F}^0,C^0)& \\
&\cdots& \\
&[\widetilde{F}^{K-1},C^{K-1}] = LSTM(F^{K-1}, \widetilde{F}^{K-2},C^{K-2}),& \\
\end{aligned}
\end{equation}
where we set $\widetilde{F}^{init}$ and $C^{init}$ to be zeros, which means that we do not known any information.

\begin{figure}[h]
\centering
    \includegraphics[width=0.9\hsize \hspace{0.01\hsize}]{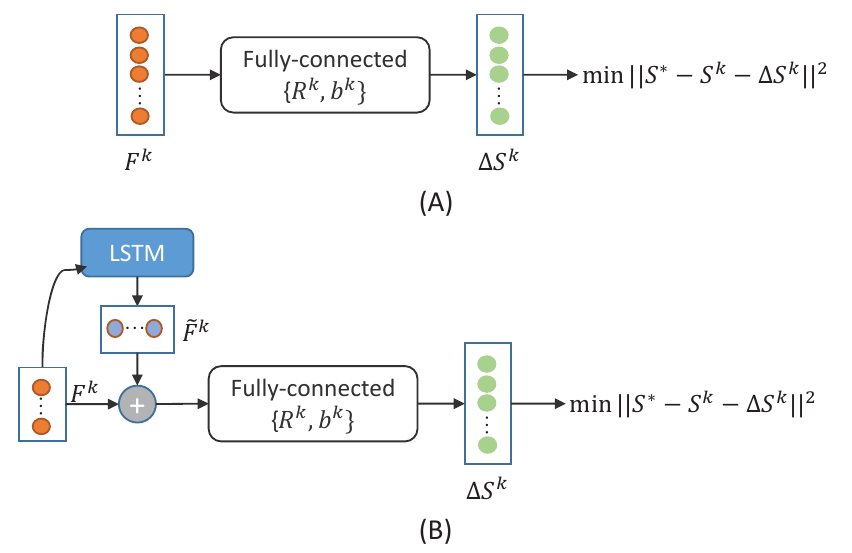}
  \caption{(A) is the traditional building block for the $k$-th iteration. (B) is our proposed building block, which a new recurrent  shape feature  $\widetilde{F}^{k}$ is added. The $R^k$ is the weight filter and $b^k$ is the bias filter in the fully-connected layer.  }
  \label{cnn_rnn}  
\end{figure}

\textbf{Regression.} \label{regression}
A series of $\{R_k, b_k\}$ are learned in the training stage, via the following loss function:
\begin{equation}
\min ||S^{*} - S^k - \Delta S^k||^2,
\label{least_squares_k}
\end{equation}
where $\Delta S^k = R^k (F^k + \widetilde{F}^{k}) + b^k$, which is the linear combination of the current shape-indexed feature $F^k$ and the recurrent features $\widetilde{F}^{k}$ plus a biased term $b^k$. The different between the traditional cascaded regression and our method is shown in Fig. \ref{cnn_rnn}.

The problem (\ref{least_squares_k}) is the well-known least square problem. This step aims to regress the $S^{k}$ to a shape $S^{k+1}$ closer to the hand-labeled landmarks $S^{*}$.

%
Since $\{R^k, b^k\}$ represents the linear matrix and the bias, it can be rewritten as the fully-connected layer, where $R^k$ is the weight filter and $b^k$ is the bias filter in the fully-connected layer. Hence, we add $K$ fully-connected layers in our network.

\subsection{Training Objectives and Optimization}
We define two kinds of losses in the proposed deep architecture: 1) a softmax loss function designed for the initial shape detection, and 2) multiple regression losses for the recurrent facial landmark detection (each LSTM component corresponds to a regression loss). In training, we minimize the sum of these losses via back propagation and stochastic gradient descent. In the following, we present the definitions and the gradient of these two kinds of losses.

\textbf{Softmax loss function}. The softmax loss function is defined by
\begin{equation}
 \min - \sum_{i=1}^p \sum_{jk} Q^i_{jk} \log( \hat{A}^i_{jk}).
\end{equation}
Denote $\psi = \sum_{i=1}^p \sum_{jk} Q^i_{jk} \log( \hat{A}^i_{jk})$, then the gradient w.r.t. $A^i_{jk}$ to be computed as
\begin{equation}
\begin{aligned}
 \frac{\partial \psi}{\partial A^i_{jk} } & = - \frac{\partial Q^i_{jk} \log(\hat{A}^i_{jk})}{\partial A^i_{jk}} - \sum_{l \neq j \&  m \neq k} \frac{\partial Q^i_{lm} \log(\hat{A}^i_{lm})}{\partial A^i_{jk}}& \\
\\
& = -Q^i_{jk}  +  \frac{\exp(A^i_{jk})}{\sum_{jk} \exp(A^i_{jk})} 	 \left( Q^i_{jk} + \sum_{l \neq j \&  m \neq k} Q^i_{jk} \right)  & \\
& =- Q^i_{jk}  + \frac{\exp(A^i_{jk})}{\sum_{jk} \exp(A^i_{jk})}& \\
& = - Q^i_{jk} + \hat{A}^i_{jk}, & \\
 \end{aligned}
\end{equation}
where the third equality follows  $\sum_{jk} Q_{jk}^i = 1$.

Thus, the gradient w.r.t. $A^i$ is $\hat{A}^i - Q^i$.

\textbf{Recurrent Regression Loss} The regression loss for the $k+1$-th iteration is defined as:
\begin{equation}
\min ||S^{*} - S^k - \Delta S^k||^2.
\label{least_squares}
\end{equation}
The gradient w.r.t. $\Delta S^k$ can be calculated by $2 \times (\Delta S^k - S^{*} + S^k)$.

\section{Experiments}
\label{experiments}

\subsection{Datasets and Data Augmentation}

We conduct extensive evaluations of the proposed method on three benchmark datasets.
\begin{itemize}
\item \textbf{LFPW}~\cite{lfpw}: The Labeled Face Parts in-the-Wild (LFPW) database contains 1,287 images downloaded from the Internet. Due to some invalid URLs, we evaluate the performance on 811 training and 224 test images provided by~\cite{300w}.

\item \textbf{HELEN}~\cite{helen}: It contains 2,330 annotated images downloaded from the Flickr. We use 2,000 images as the training set and 330 images as testing.

\item \textbf{300-W}~\cite{300w}: The 300-W dataset consists of 3,148 training images from the LFPW, HELEN and the whole AFW~\cite{zhu2012face}. It performs testing on three parts: common subset, challenging subset and the full set. The common subset contains 554 images from LFPW and HELEN databases and the challenging subset contains 135 images from IBUG. The full set is the union of them (689 images).
\end{itemize}

We conduct evaluations on 68 points (provided by \cite{300w}) on the LFPW, HELEN and 300-W  datasets.

\textbf{Data augmentation.}
We train our models only using the data from the training data without external sources. To reduce overfitting on the training data, we employ three distinct forms of data augmentation to artificially enlarge the dataset.

The first form of data augmentation is to generate image rotations. We do this by rotating the image into different angles including $\{\pm 30,\pm 25, \pm 20, \pm 15, \pm 10, \pm 5, 0\}$.

The second form of data augmentation is to disturb the bounding boxes, which can increase the robustness of our results to the bounding boxes. We randomly scale and translate the bounding box for each image.

The third form of data augmentation is mirroring. We flip all images and their shapes.

After the data augmentation, the number of training samples is enlarged to 52 times, which is shown in Table~\ref{number_training}.

\begin{table}[h]
\small
    \centering \caption{The number of training samples after data augmentation.}
    \begin{tabular}{|c|c|c|}
         \hline
          LFPW & HELEN & 300-W \\
        \hline
           42,172 & 121,160  & 163,696\\
         \hline
        \end{tabular}
        \label{number_training}
\end{table}

\subsection{Experimental Setting}

\textbf{Implementation details.} We implement the proposed method based on the open source Caffe~\cite{jia2013caffe} framework, which is an efficient deep neural network implementation. Note that the Caffe also includes the implement of the LSTM layer \footnote{https://github.com/LisaAnne/lisa-caffe-public/tree/lstm\_video\_deploy}. We first crop the image using the bounding box with the  $0.2 W$ padding on all sides (top, bottom, left, right), where $W$ is the width of the bounding box. We scale the longest side to 256 leaving us with a $256 \times H$ or $H \times 256$ sized image, where $H \leq 256$. Then we add zeros to the smallest side and make the size of $256 \times 256$ pixels. The number of candidate shapes is set to $N=50$. We set $K=8$ and $b = 6$. Our network is trained by stochastic gradient descent with 0.9 momentum. The weight decay parameter is 0.0001. The network's parameters are initialized with the pre-trained VGG19 model.

\textbf{Evaluation.}
We evaluate the alignment accuracies by two popular metrics, the mean error and the cumulative errors.
The mean error is measured by the distances between the predicted landmarks and the ground truths, normalized by the inter-pupil distance,
 which can be calculated by
\begin{equation}
\text{mean \ error} = \frac{1}{n} \sum_{i=1}^n \frac{ ||S_i - S^{*}_i||^2}{pD_i}
\end{equation}
where $S_i$ is the predicted shape and $S_i^{*}$ is the ground-truth shape for the $i$-th image. $D_i$ is the distance between two eyes. $p$ is the number of landmarks and $n$ is the total number of face images.

We also report the cumulative errors distribution (CED) curve, in which the mean error larger than $l$ is reported as a failure. Let $e_i =\frac{ ||S_i - S^{*}_i||^2}{pD_i}$, and CED at the error $l$ is defined as
\begin{equation}
\text{CED} = \frac{N_{e \leq l}}{n},
\end{equation}
where $N_{e \leq l}$ is the number of images on which the error $e_i$ is no higher than $l$.

\subsection{Comparison with State-of-the-art Algorithms}
The first set of experiments is to evaluate the performance of the proposed method and compare it with several state-of-the-art algorithms. Zhu et al.~\cite{zhu2012face},  DRMF~\cite{asthana2013robust}, ESR~\cite{cao2014face}, RCPR~\cite{burgos2013robust}, SDM~\cite{sdm}, Smith et al.~\cite{smith2014nonparametric}, Zhao et al.~\cite{zhao2014unified}, GN-DPM~\cite{tzimiropoulos2014gauss}, CFAN~\cite{zhang2014coarse}, ERT~\cite{kazemi2014one}, LBF~\cite{lbf}, cGPRT~\cite{lee2015face}, CFSS~\cite{cfss} and TCDCN~\cite{zhang2015learning} are selected as the baselines.

\subsubsection{Comparison on LFPW}
The goal of LFPW is to evaluate the facial landmark detection algorithms under unconstrained conditions. The images include different poses, expressions, illuminations and occlusions, and mainly are collected from the Internet.

\begin{table}[h]
\small
    \centering \caption{Mean Error on LFPW dataset.}
    \begin{tabular}{c c }
        \hline
 \multicolumn{2}{c}{LFPW Dataset} \\
Methods & 68 pts \\
        \hline
 Zhu et al. & 8.29   \\

 DRMF  & 6.57 \\

 RCPR & 5.67 \\

 SDM & 5.67  \\

 GN-DPM & 5.92 \\

 CFAN & 5.44 \\

 CFSS & 4.87 \\

 CFSS Practical & 4.90   \\
 \hline

 OURS & \textbf{4.49} \\
 \hline
        \end{tabular}
    \label{lfpw}
\end{table}

We compare the proposed method with the several state-of-the-art methods as shown in Table \ref{lfpw}. As can be seen, the proposed deep-network based facial landmark detection is significantly better than the baselines. The most similar work is SDM which is also a cascaded regression method. The main different is that it is trained by the hand-crafted features. The mean error of SDM is 5.67, while the proposed method is 4.49. CFSS is the recent proposed method, which achieves an excellent performance on this dataset. Even so, our method also performances better that CFSS, and shows an error reduction of 0.38.

\subsubsection{Comparison on HELEN}
The images of HELEN are downloaded from Flickr, which under a broad range of appearance variation, including lighting, individual differences, occlusion and pose. The HELEN dataset contains many sufficiently large faces (greater than 500 pixels in width) and can fit accurately for high resolution images.

Table \ref{helen} shows the comparison results of mean error on the HELEN dataset. The comparison shows our proposed method can achieve better results than the baselines. For example, the mean error of our method is 4.02, compared to the 4.60 of the second best algorithm.

\begin{table}[h]
\small
    \centering \caption{Mean Error on HELEN database.}
    \begin{tabular}{c c }
        \hline
\multicolumn{2}{c}{Helen Dataset} \\
Methods & 68 pts \\
        \hline
  Zhu et al. & 8.16  \\

  DRMF & 6.70  \\

  ESR & 5.70   \\

  RCPR & 5.93  \\

  SDM & 5.50  \\

  GN-DPM & 5.69 \\

 CFAN & 5.53  \\

 CFSS& 4.63  \\

  CFSS Practical  & 4.72  \\
  TCDCN  & 4.60  \\
 \hline
  OURS & \textbf{4.02 }\\
 \hline
        \end{tabular}
    \label{helen}
\end{table}

\subsubsection{Comparison on 300W}
The 300W is extremely challenging dataset, which is widely used for comparing the performance of different algorithms of facial landmark detection under the same evaluation protocol.

\begin{table}[h]
\small
    \centering \caption{Mean Error on 300W database. }
    \begin{tabular}{c c c c}
        \hline
  \multicolumn{4}{c}{300-W Dataset}\\
 Methods & Common & Challenging & Fullset \\
        \hline
 Zhu et al. & 8.22 & 18.33 & 10.20 \\

  DRMF & 6.65 & 19.79 & 9.22 \\

  ESR & 5.28 & 17.00 & 7.58  \\

  RCPR & 6.18 & 17.26 & 8.35 \\

  SDM & 5.57 & 15.40 & 7.50 \\
  Smith et al. & & 13.30 & \\

  Zhao et al. & &  & 6.31 \\

 GN-DPM & 5.78 & &\\

 CFAN  & 5.50 & & \\

 ERT& & & 6.40 \\

 LBF  & 4.95 & 11.98 & 6.32 \\

 LBF fast  & 5.38 & 15.50 & 7.37 \\

 cGPRT  &  &  & 5.71 \\

 CFSS & 4.73 & 9.98 & 5.76 \\

  CFSS Practical & 4.79 & 10.92 & 5.99 \\
  TCDCN & 4.80 & 8.60 & 5.54 \\
 \hline

 OURS & \textbf{4.07} &  \textbf{8.29} & \textbf{4.90}\\
 \hline
        \end{tabular}
    \label{mean_error_300w}
\end{table}

Table \ref{mean_error_300w} shows the comparison results of mean error on the 300W dataset. It can be observed that the proposed method performs significantly better than all previous methods in all settings. Specifically, on Fullset, our method obtains a mean error of 4.90, which gives an error reduction of $0.64$ compared to the second best algorithm. On Challenging, our method shows an error reduction of $0.31$ in comparison with the second best method. On Common, the mean error of our method is 4.07, compared to 4.73 of the second best algorithm.

\begin{figure}
\centering
    \includegraphics[width=0.9\hsize \hspace{0.01\hsize}]{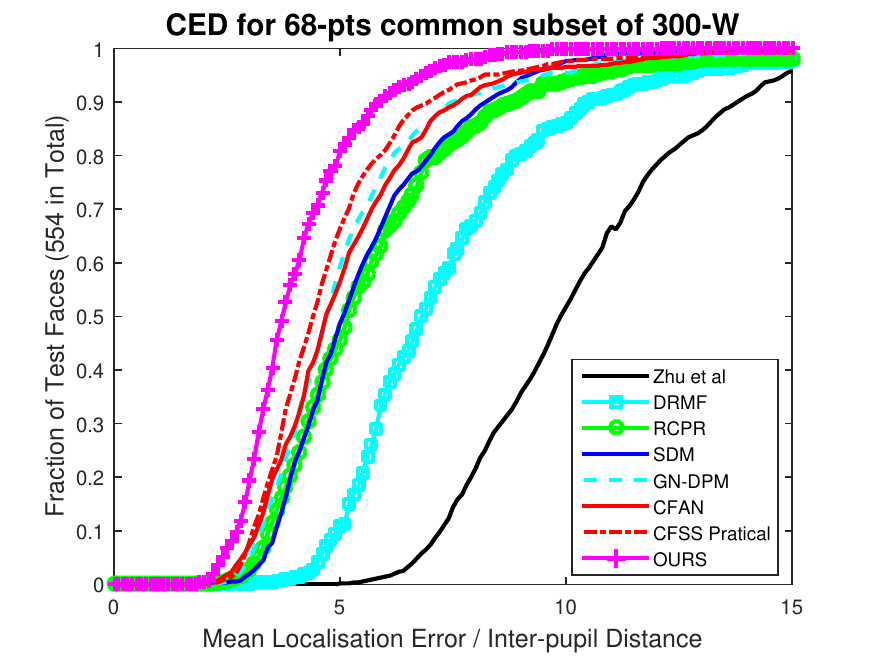}
  \caption{CED curves on common subset of 300-W.}
  \label{ced_common}  
\end{figure}

\begin{figure}
\centering
    \includegraphics[width=0.9\hsize \hspace{0.01\hsize}]{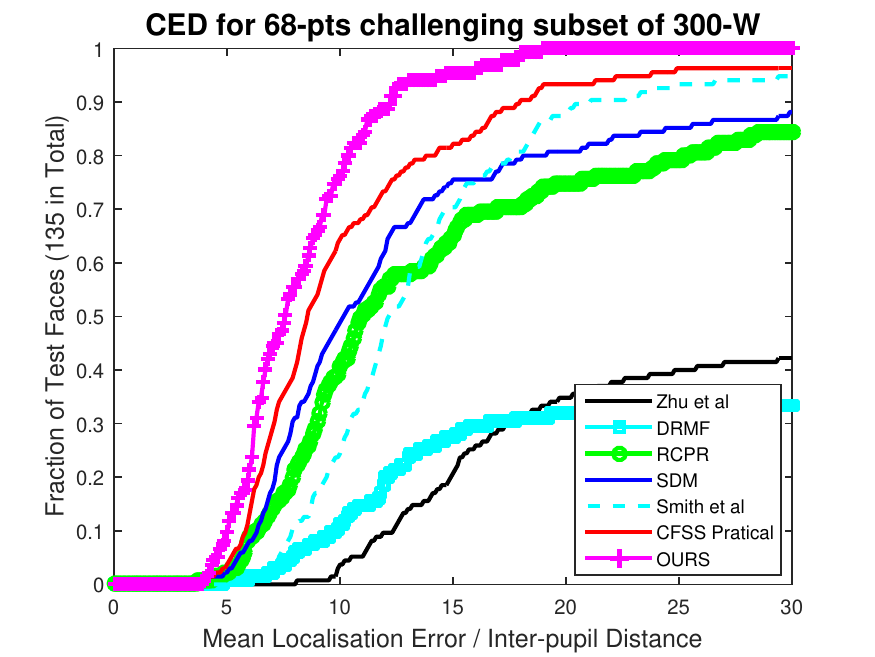}
  \caption{CED curves on challenging subset of 300-W.}
  \label{ced_challenge}  
\end{figure}
Fig.~\ref{ced_common} shows the CED curves for different error levels on the 300-W dataset. Again, for all error levels, our method yields the highest accuracy and beats all the baselines.
For instance, the proposed method shows a relative increase of $23\%$ on the 300w common set compared to the second best algorithm. The example alignment results of our method are shown in Table \ref{image_examples_common} and Table \ref{image_examples}.

\begin{table*}[t]
 \centering \caption{Shape detection examples from 300W Common Subset.}
\begin{tabular}{|c|c|c|c|c|c|c|c|c|}
\hline
\includegraphics[scale=0.09]{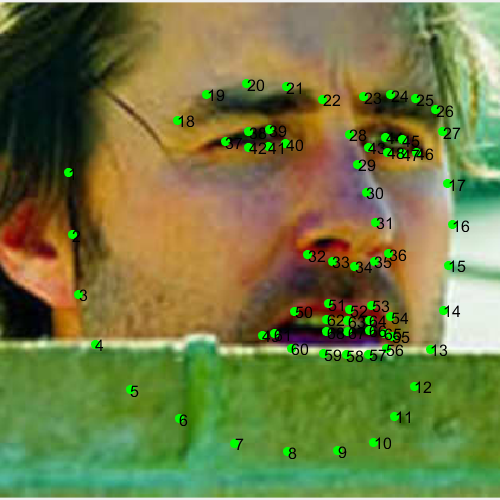}&
\includegraphics[scale=0.09]{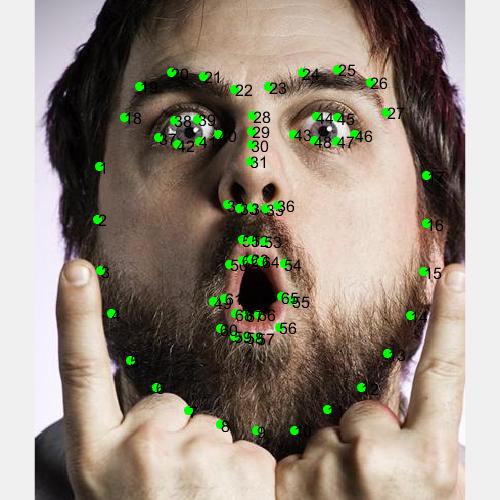}&
\includegraphics[scale=0.09]{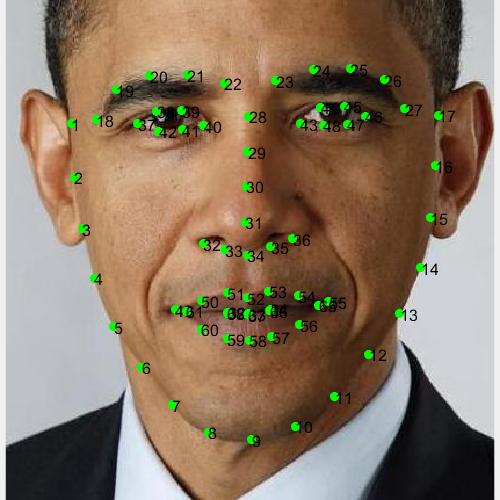} &
\includegraphics[scale=0.09]{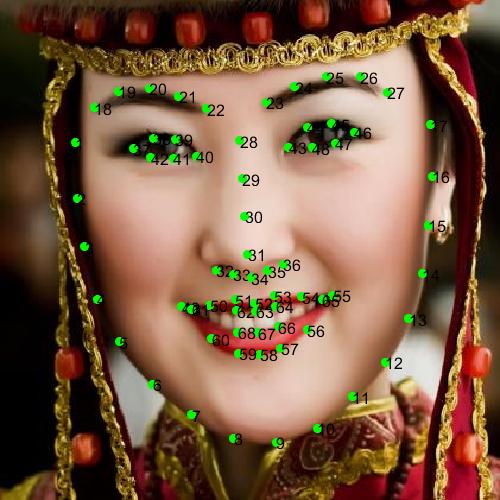} &
\includegraphics[scale=0.09]{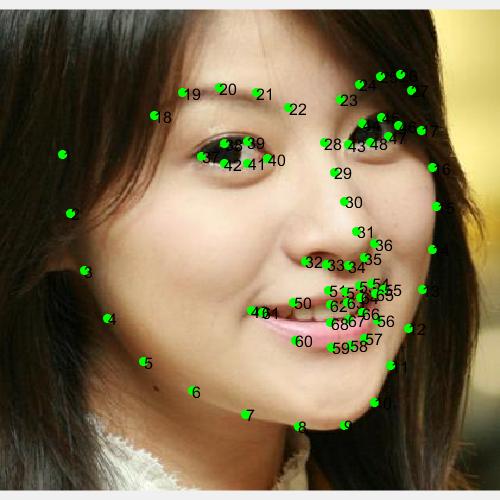} &
\includegraphics[scale=0.09]{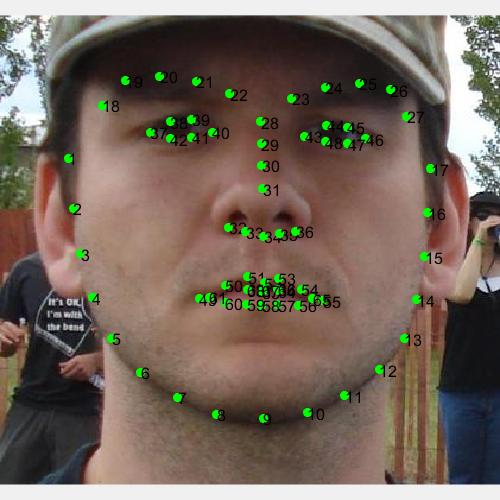}&
\includegraphics[scale=0.09]{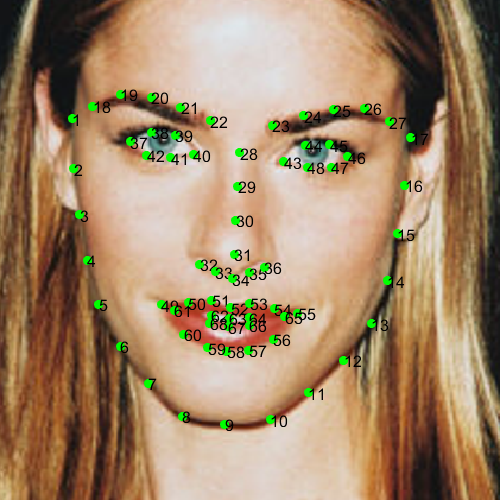}&
\includegraphics[scale=0.09]{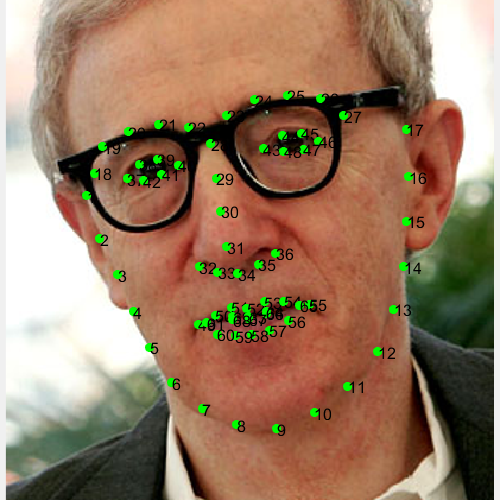} &
\includegraphics[scale=0.09]{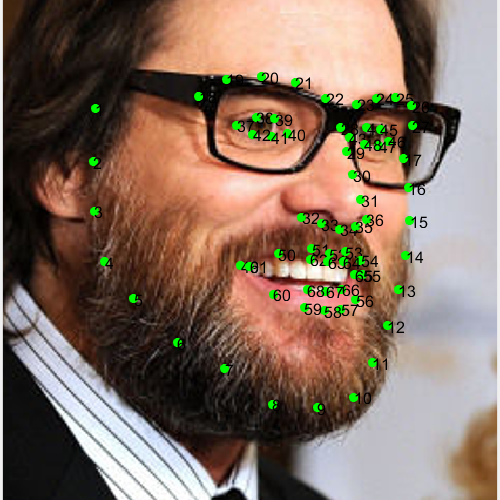} \\
\hline
\includegraphics[scale=0.09]{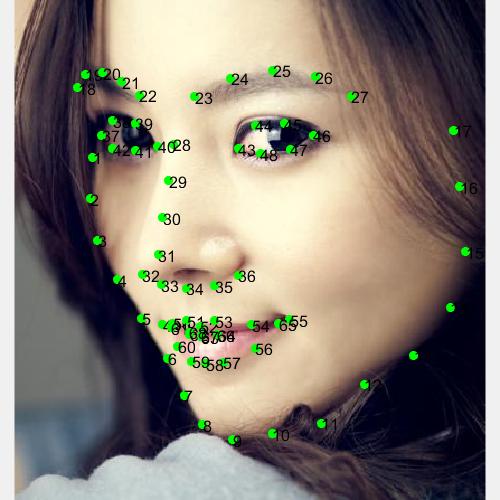}&
\includegraphics[scale=0.09]{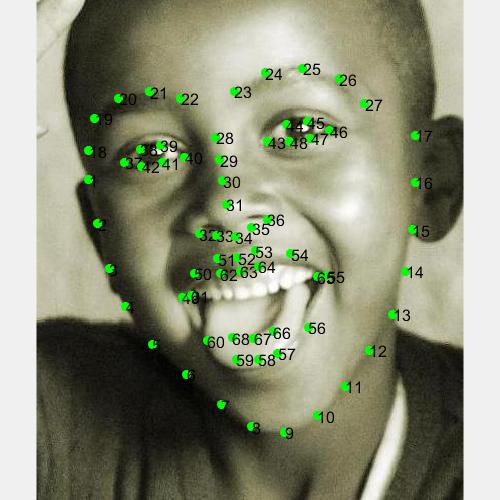}&
\includegraphics[scale=0.09]{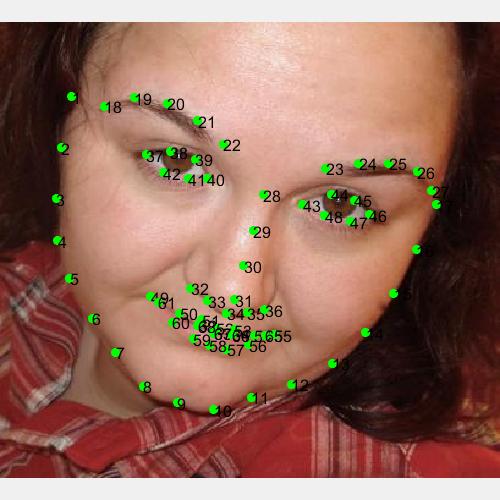} &
\includegraphics[scale=0.09]{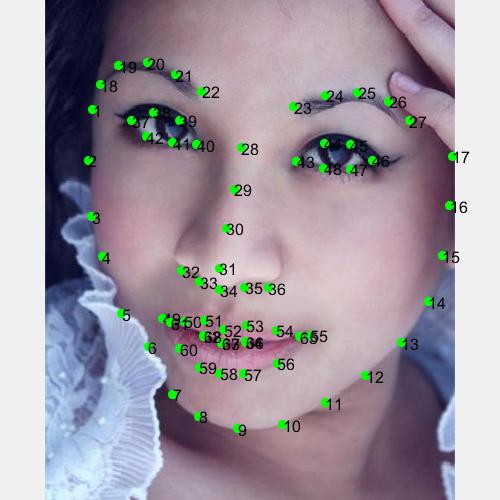} &
\includegraphics[scale=0.09]{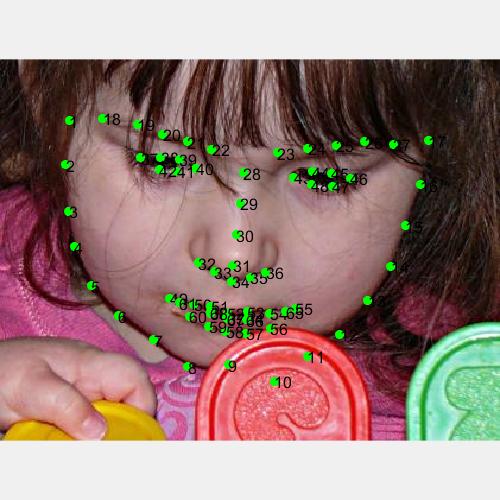} &
\includegraphics[scale=0.09]{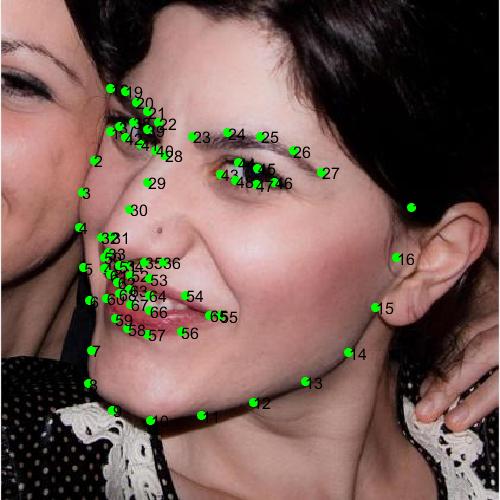}&
\includegraphics[scale=0.09]{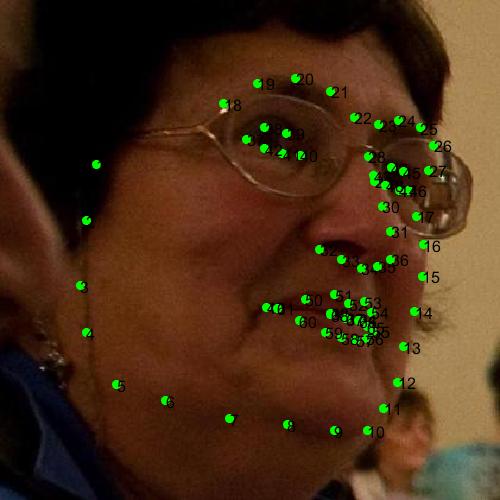}&
\includegraphics[scale=0.09]{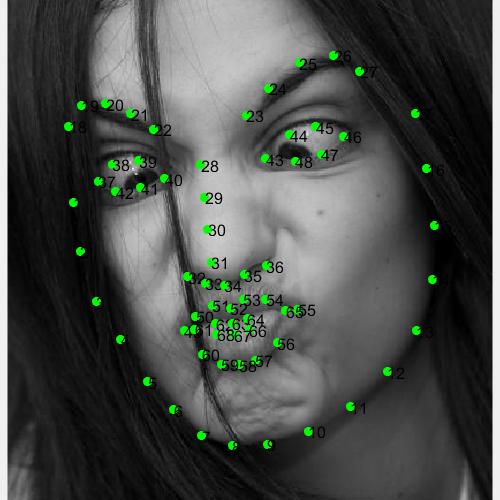} &
\includegraphics[scale=0.09]{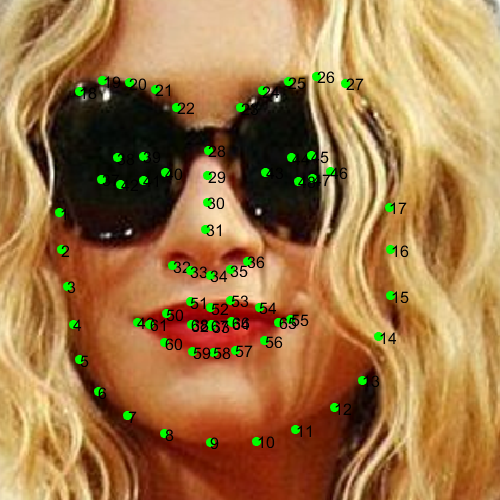} \\
\hline
\includegraphics[scale=0.09]{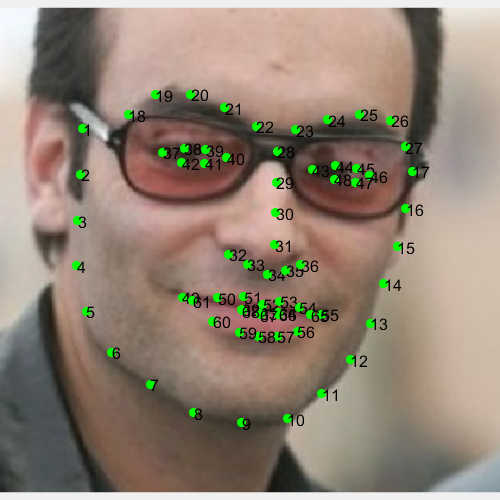}&
\includegraphics[scale=0.09]{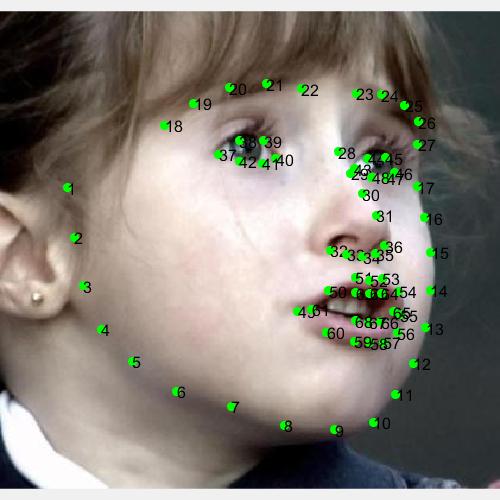}&
\includegraphics[scale=0.09]{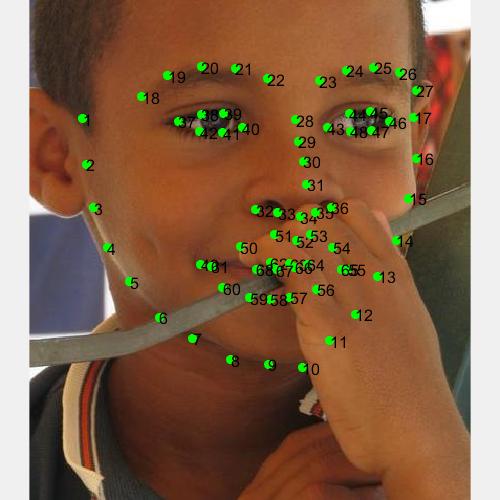} &
\includegraphics[scale=0.09]{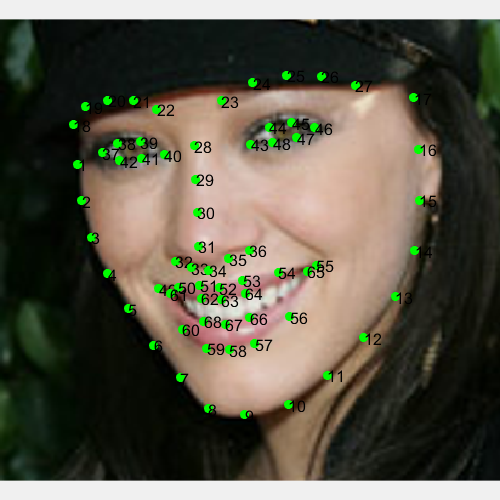} &
\includegraphics[scale=0.09]{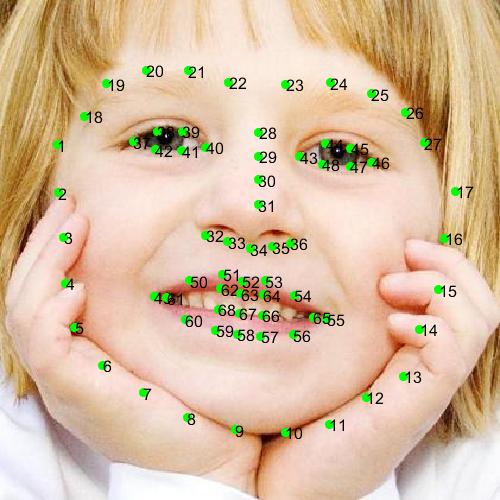} &
\includegraphics[scale=0.09]{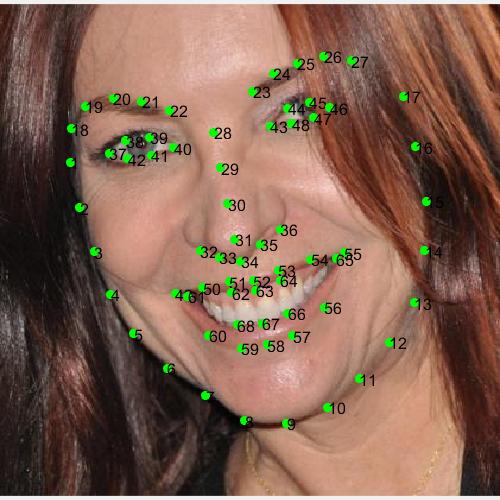}&
\includegraphics[scale=0.09]{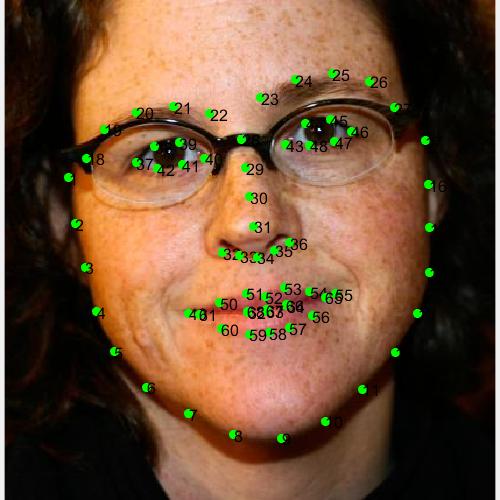}&
\includegraphics[scale=0.09]{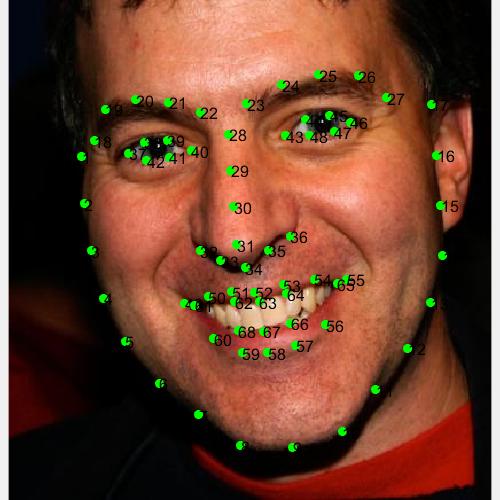} &
\includegraphics[scale=0.09]{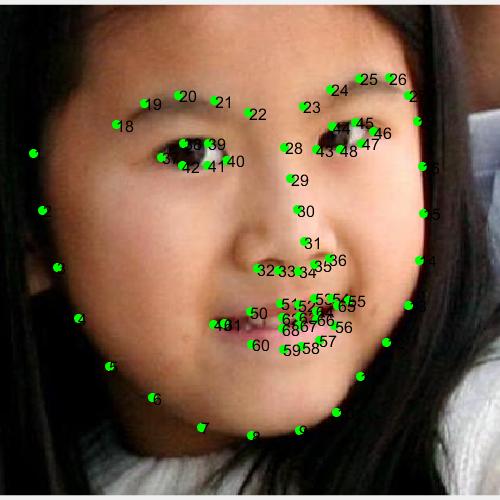} \\
\hline
\includegraphics[scale=0.09]{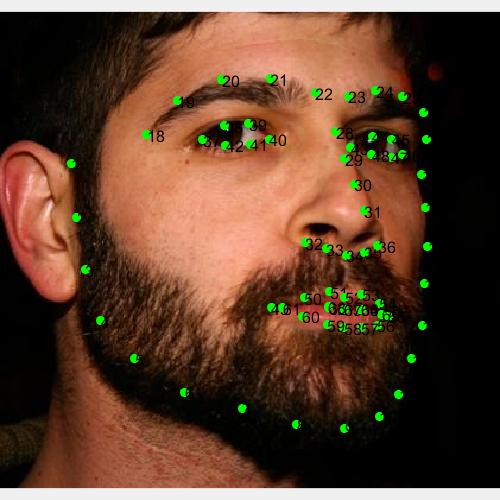}&
\includegraphics[scale=0.09]{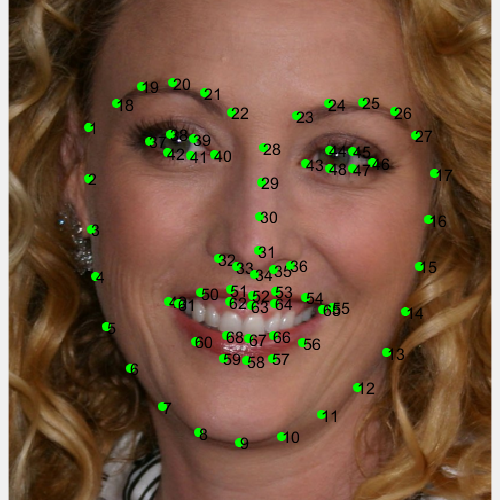}&
\includegraphics[scale=0.09]{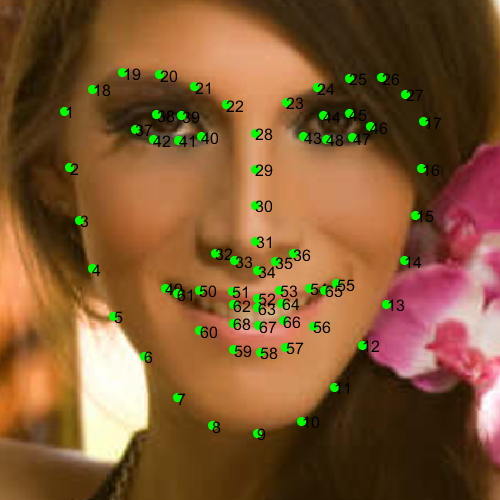} &
\includegraphics[scale=0.09]{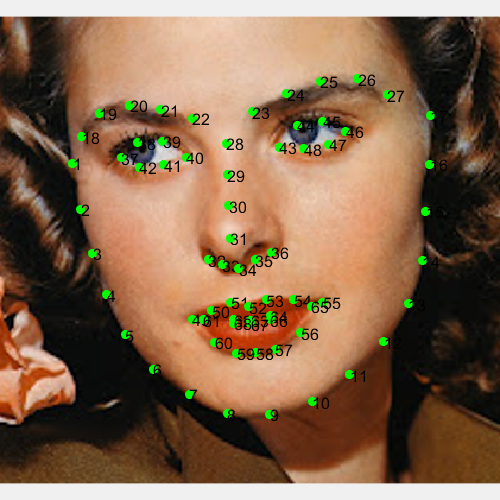} &
\includegraphics[scale=0.09]{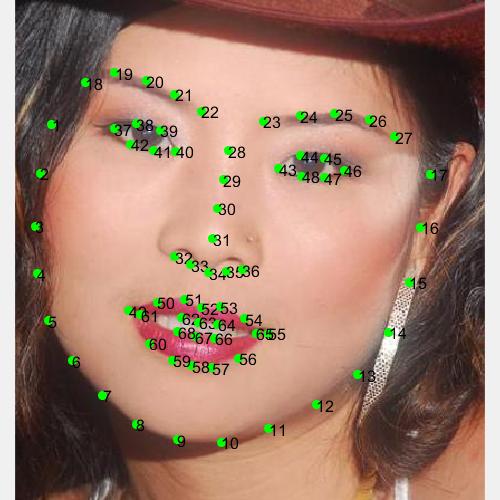} &
\includegraphics[scale=0.09]{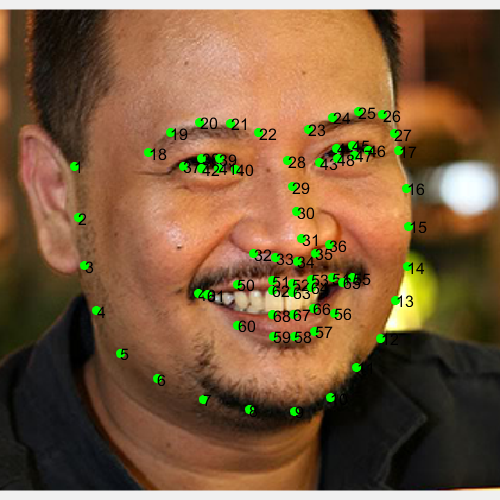}&
\includegraphics[scale=0.09]{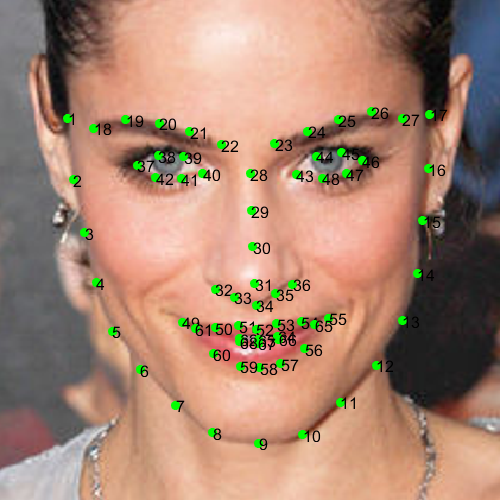}&
\includegraphics[scale=0.09]{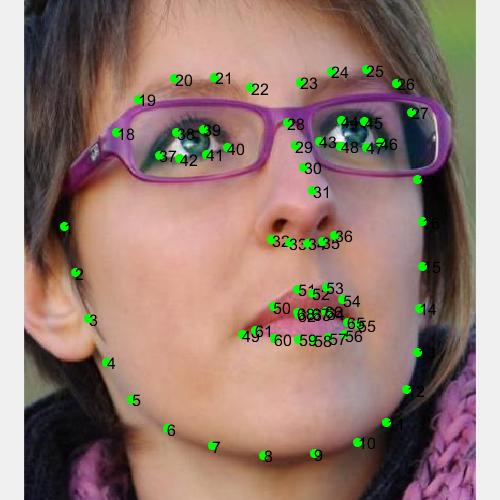} &
\includegraphics[scale=0.09]{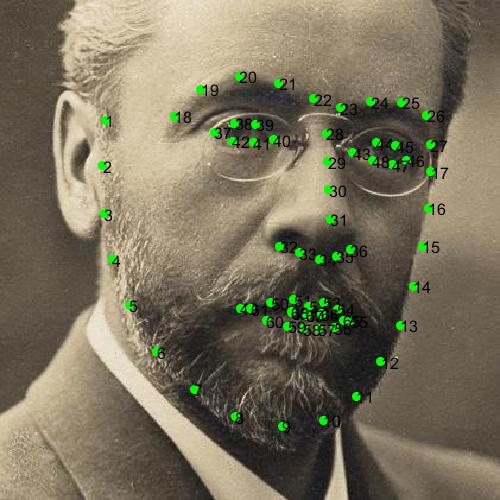} \\
\end{tabular}
 \label{image_examples_common}
\end{table*}

\begin{table*}[t]
 \centering \caption{Shape detection examples from 300W Challenge Subset.}
\begin{tabular}{|c|c|c|c|c|c|c|c|c|}
\hline
\includegraphics[scale=0.09]{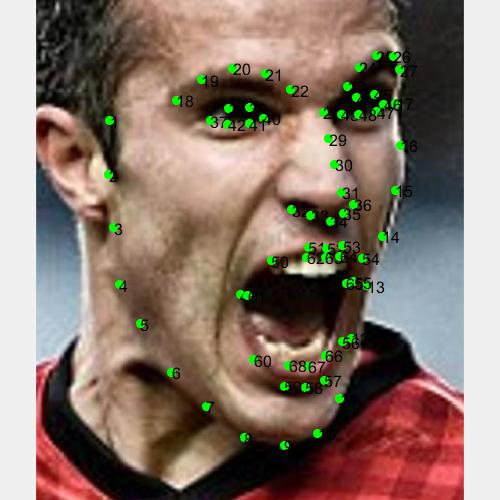}&
\includegraphics[scale=0.09]{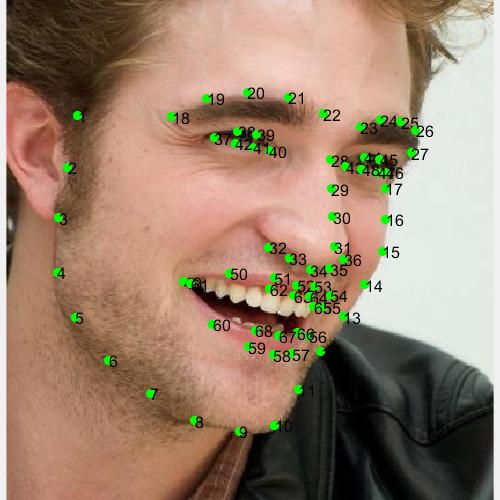}&
\includegraphics[scale=0.09]{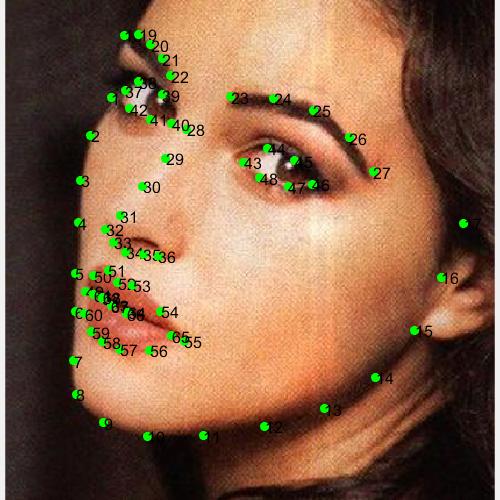} &
\includegraphics[scale=0.09]{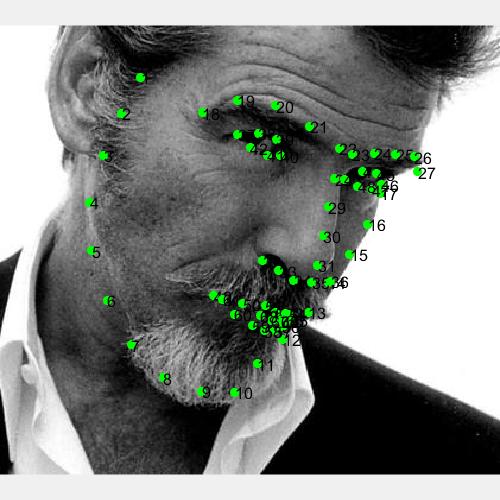} &
\includegraphics[scale=0.09]{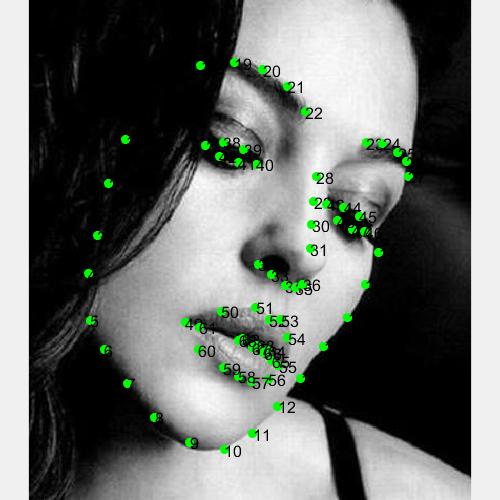} &
\includegraphics[scale=0.09]{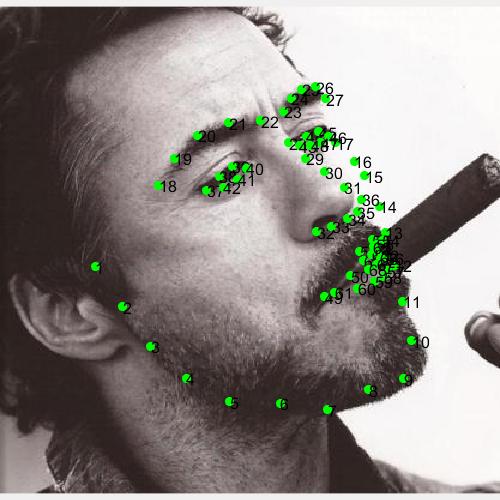}&
\includegraphics[scale=0.09]{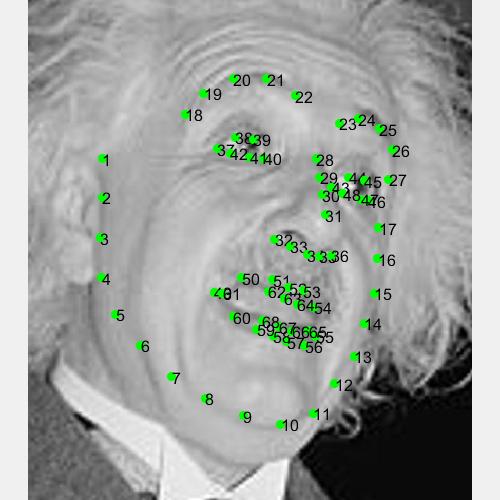}&
\includegraphics[scale=0.09]{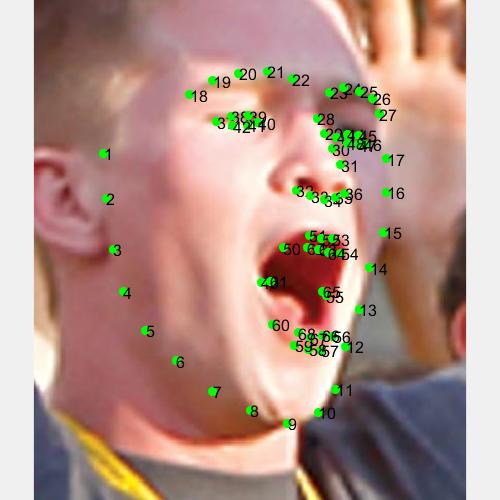} &
\includegraphics[scale=0.09]{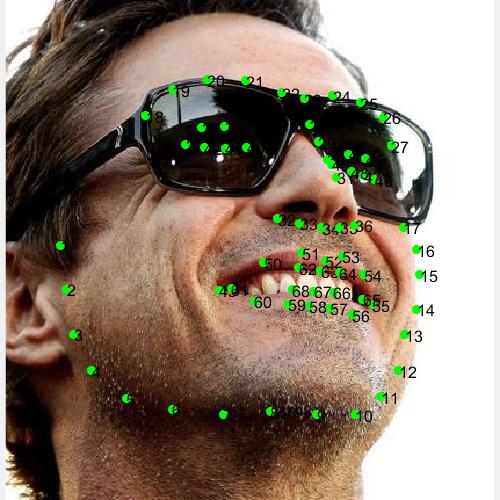} \\
\hline
\includegraphics[scale=0.09]{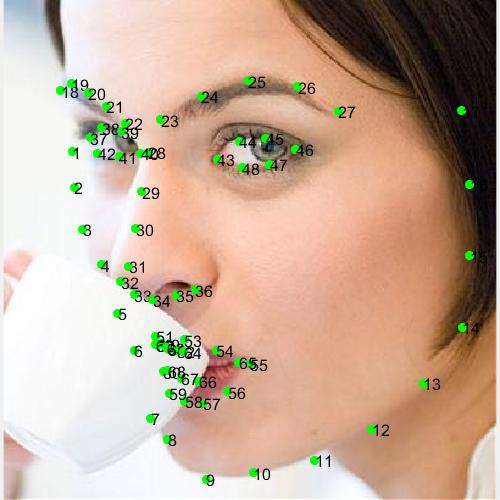} &
\includegraphics[scale=0.09]{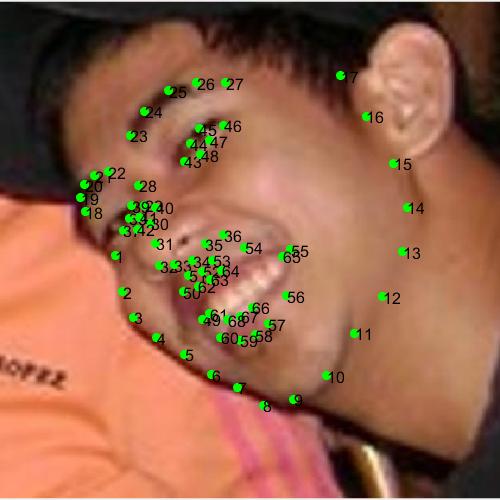}&
\includegraphics[scale=0.09]{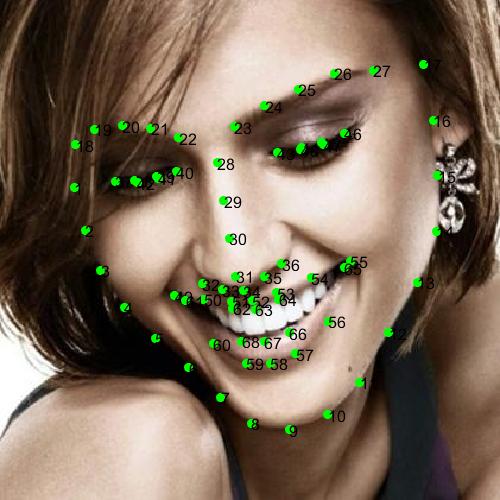}&
\includegraphics[scale=0.09]{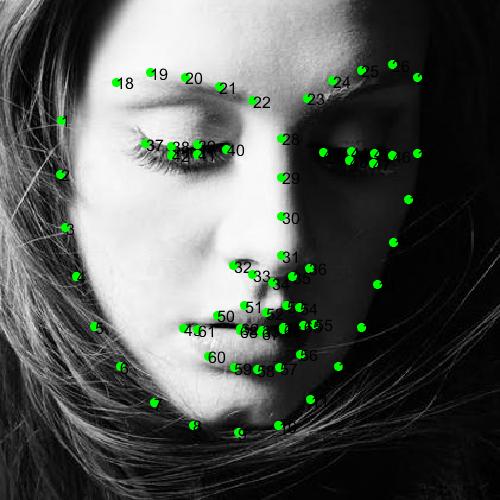} &
\includegraphics[scale=0.09]{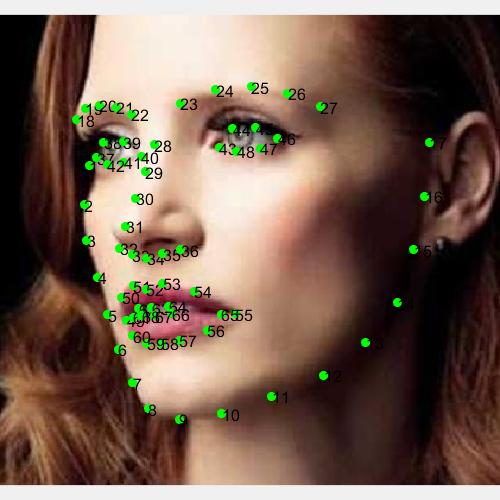} &
\includegraphics[scale=0.09]{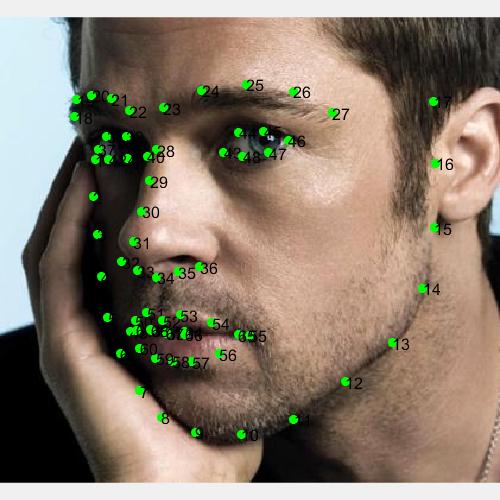} &
\includegraphics[scale=0.09]{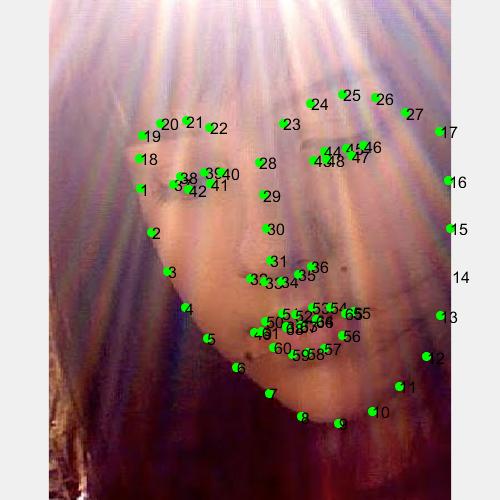}&
\includegraphics[scale=0.09]{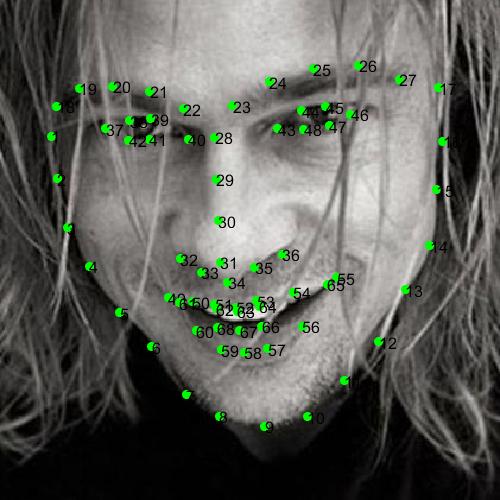}&
\includegraphics[scale=0.09]{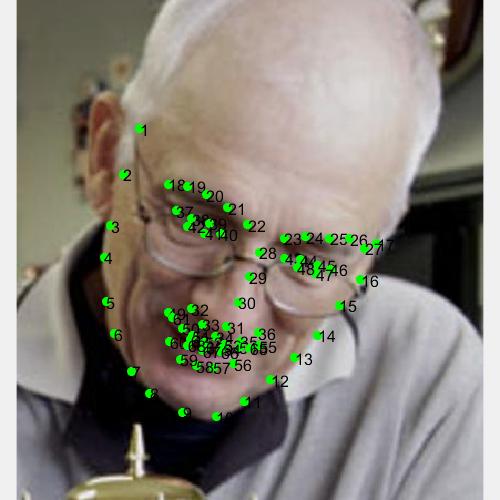} \\
\hline
\includegraphics[scale=0.09]{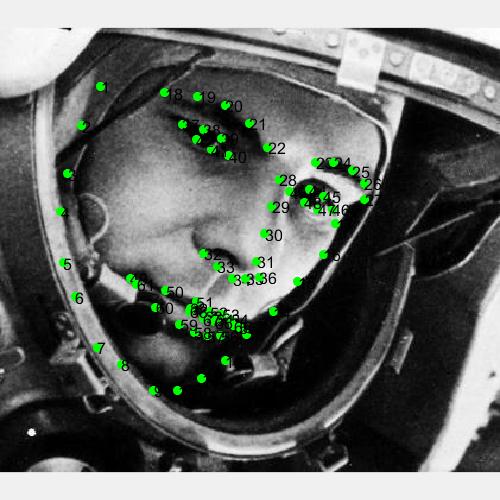} &
\includegraphics[scale=0.09]{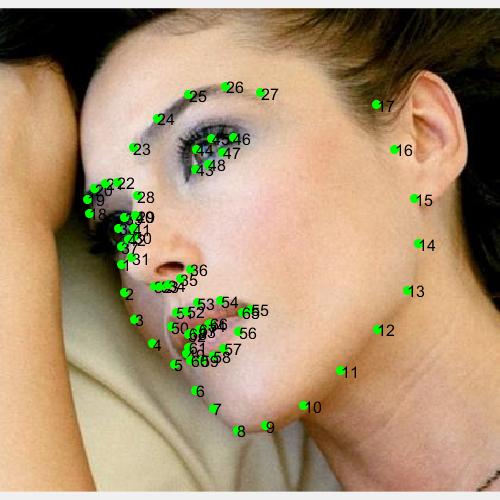} &
\includegraphics[scale=0.09]{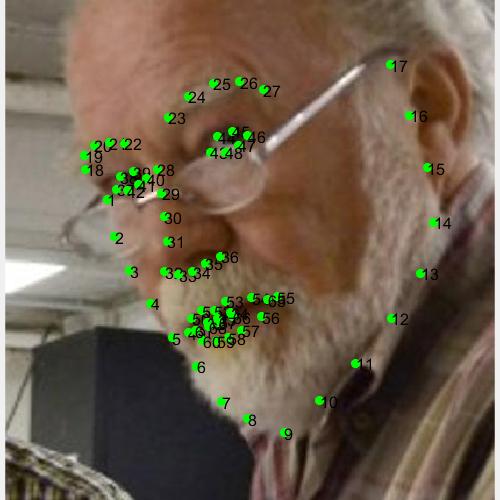}&
\includegraphics[scale=0.09]{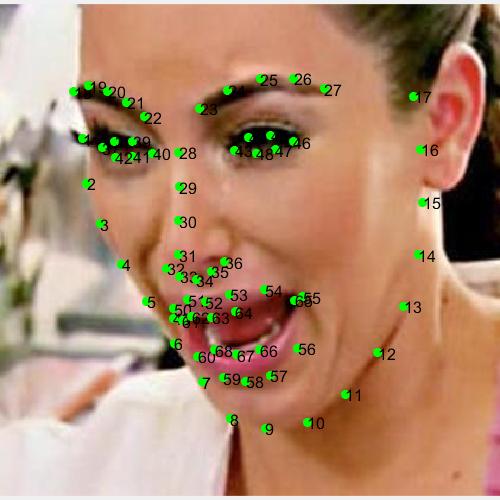}&
\includegraphics[scale=0.09]{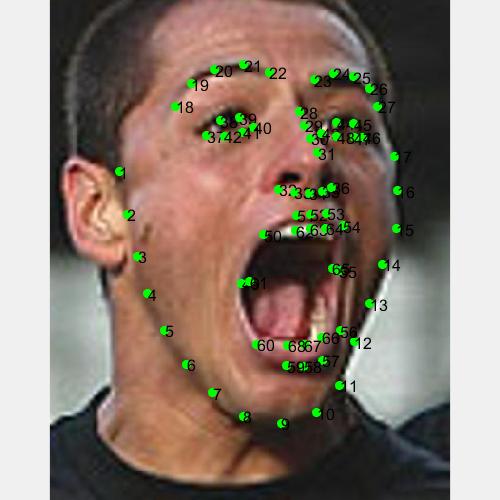} &
\includegraphics[scale=0.09]{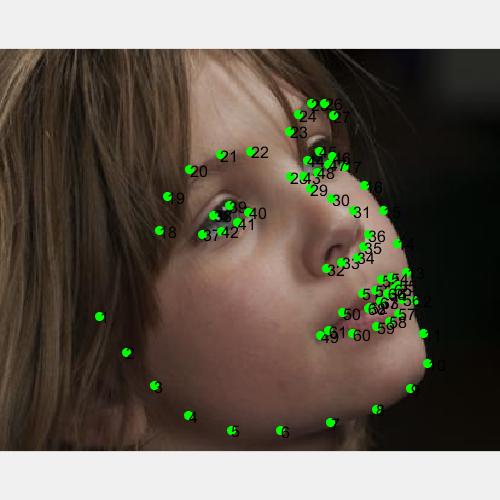} &
\includegraphics[scale=0.09]{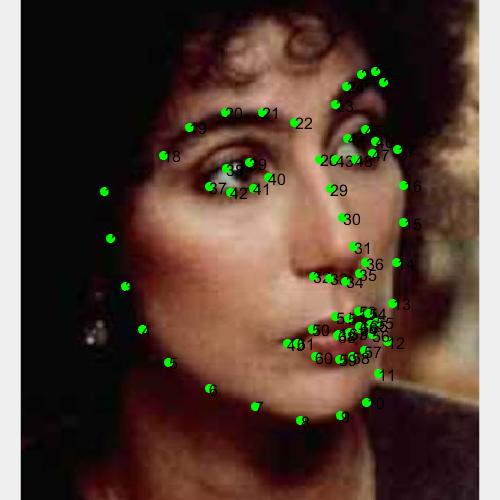} &
\includegraphics[scale=0.09]{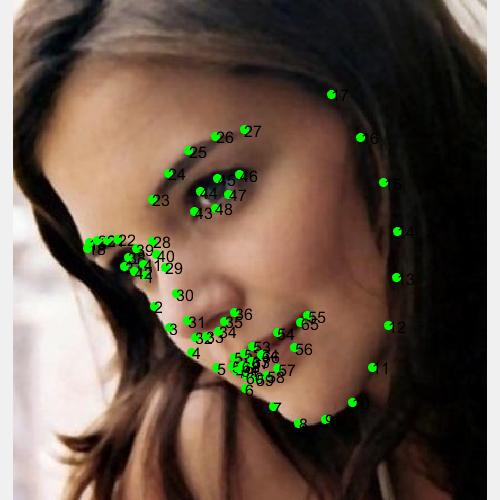}&
\includegraphics[scale=0.09]{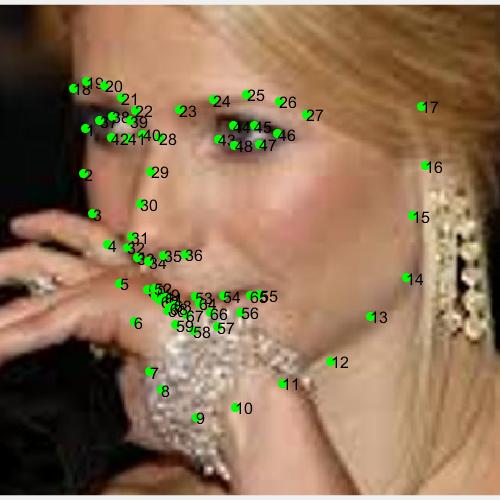} \\
\hline
\includegraphics[scale=0.09]{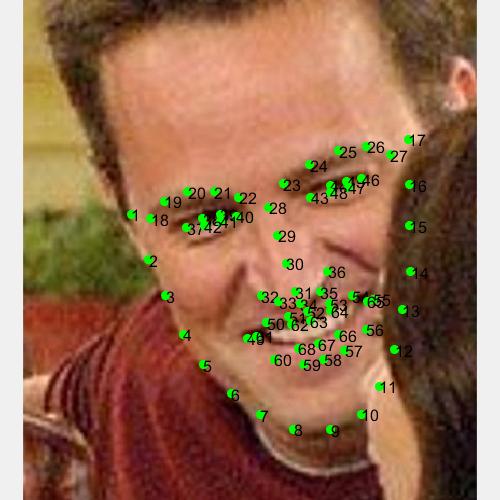} &
\includegraphics[scale=0.09]{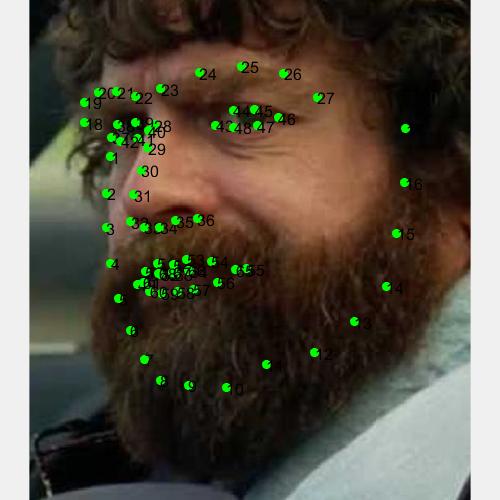} &
\includegraphics[scale=0.09]{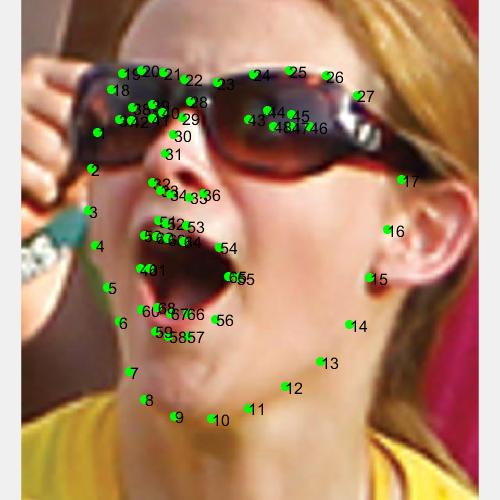}&
\includegraphics[scale=0.09]{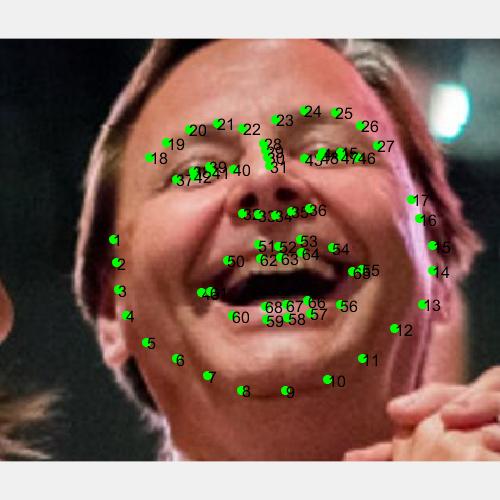}&
\includegraphics[scale=0.09]{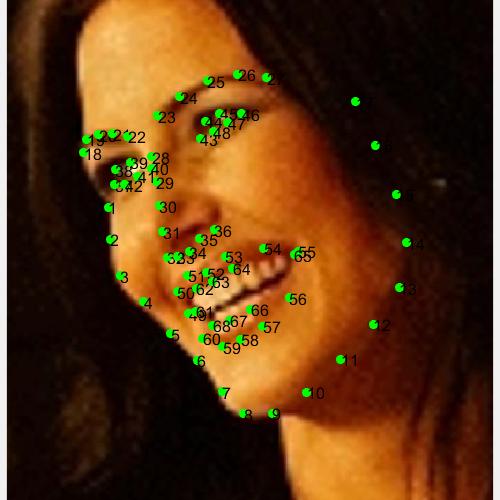} &
\includegraphics[scale=0.09]{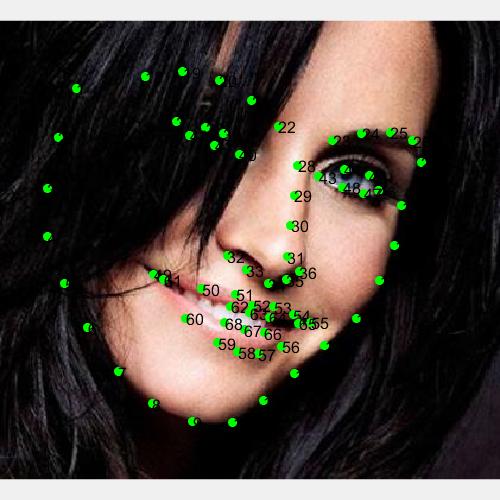} &
\includegraphics[scale=0.09]{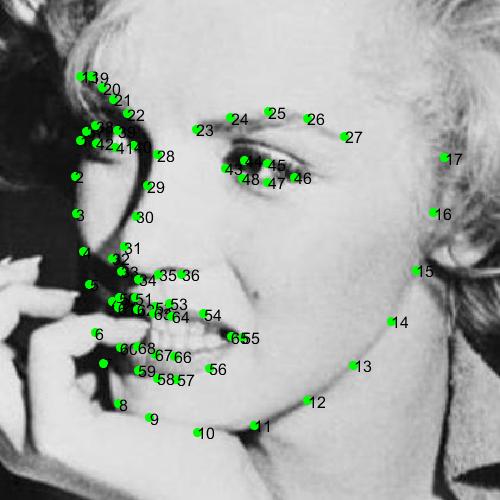} &
\includegraphics[scale=0.09]{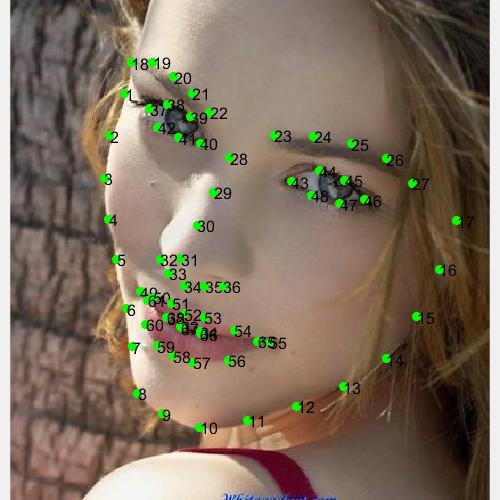}&
\includegraphics[scale=0.09]{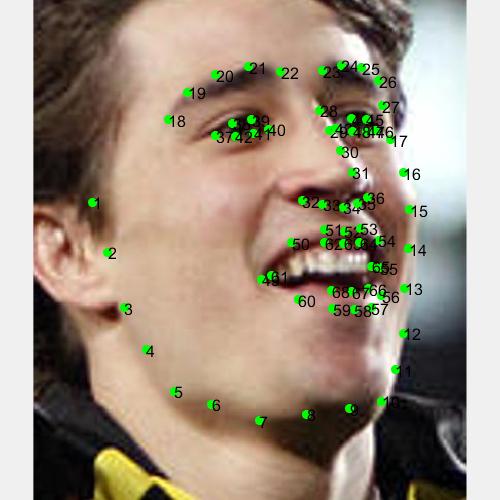} \\
\hline
\end{tabular}
 \label{image_examples}
\end{table*}

One main reason for the good performance of our method is that instead of using traditional hand-crafted visual features (SIFT, HOG), it uses the deep network to learn the image representations and extracts the deep shape-indexed features.
Second, our method can incorporate the previous shape information to the current shape regression, which can help to obtain a good performance.

\subsection{Further Analyses}

\subsubsection{Effects of Different Input Sizes and Networks}
Some may argue that our network is too large and deep, it may impractical to be used. Since the most time consuming module is the spatial resolution-preserved network, we explore the effects of different sub-networks in this subsection. Also the input size of image can effect the running time, e.g., the running time of $128 \times 128$'s image is 4 times faster than that of $256 \times 256$'s. Hence, we also explore the effects of different sizes of input images.

\begin{table}[t]
    \centering \caption{Configurations of the Spatial Resolution-Preserved Conv-DeConv Network for VGG-S with the input size : $256 \times 256$.}
    \begin{tabular}{|c|c | c |}
        \hline
           {\bf type} &  {\bf filter size/stride/pad } & {\bf output size} \\
        \hline
         conv1 & 7 $\times$ 7 / 1 / 3 & 96 $\times$ 256 $\times$ 256 \\
           \hline
        max pool1 & 2 $\times$ 2 / 2 / 0  & 96 $\times$ 128 $\times$ 128 \\
         \hline
         conv2 & 5$\times$ 5 / 1 / 2  & 256 $\times$ 128 $\times$ 128 \\
         \hline
        max pool2 & 2 $\times$ 2 / 2 / 0  & 256 $\times$ 64 $\times$ 64 \\
         \hline
         conv3 & 3$\times$ 3 / 1 / 1  & 512 $\times$ 64 $\times$ 64 \\
         \hline
         conv4 & 3$\times$ 3 / 1 / 1  & 512 $\times$ 64 $\times$ 64 \\
         \hline
         conv5 & 3$\times$ 3 / 1 / 1  & 512 $\times$ 64 $\times$ 64 \\
         \hline
         deconv6 & 4$\times$ 4 / 2 / 1 & 96 $\times$ 128 $\times$ 128 \\
         \hline
         deconv7 & 4$\times$ 4 / 2 / 1 & 96 $\times$ 256 $\times$ 256 \\
         \hline
        \end{tabular}
    \label{Network_def}
\end{table}

We show the results of two different types of frameworks: VGG-S~\footnote{https://github.com/BVLC/caffe/wiki/Model-Zoo} and VGG-19, where VGG-S is a small convolutional neural network and it only has five convolutional layers. Note that same modifications are made on the two frameworks as described above (i.e., removing all the fully-connected layers, keeping the first two pooling layers, other pooling layers are removed and adding two deconvolutional layers).  After the modification, it has five convolutional layers, two max pooling layers and two deconvolutional layers. Table.~\ref{Network_def} shows the network architecture of VGG-S. Compared to VGG-19, VGG-S is a very small network. This set of experiments is to show us that whether our method can performs well in the small network. We also report the results with two  different input sizes:  $256 \times 256$ and $128 \times 128$.

Table \ref{different_size} shows the comparison results, from which it can be seen that: our method can perform well even using very small sub-network. For example, the mean error of fullset is 4.90 when the sub-network is VGG-19, compared to 5.01 when the sub-network is VGG-S. Note that the mean error of TCDCN is 5.54, which is the second best results on Table \ref{mean_error_300w}. The mean error of challenging is 8.29 when the input size is $256 \times 256$, compared to 8.80 when the input size is $128 \times 128$. The results are not surprising since it is hard to discriminate the key facial points in the small images. The results also show that keeping the spatial information is important. In summary,  our framework is capable of exploiting different types of characteristics, i.e., accuracy or speed, by using different sub-networks or different input sizes.

\begin{table}[h]
\small
    \centering \caption{Mean Errors on different networks and different input sizes.}
    \begin{tabular}{c|c c c}
      \hline
 & \multicolumn{3}{|c}{300-W}\\
    & Common & Challenging & Fullset \\
        \hline
        VGG-S (128x128) & 4.62 & 9.01 & 5.48 \\
        VGG-S (256x256) & 4.22 & 8.23 & 5.01 \\
        \hline
        VGG-19 (128x128) & 4.57 & 8.80 & 5.40 \\
        VGG-19 (256x256) & 4.07 & 8.29 & 4.90\\
        \hline
        \end{tabular}
    \label{different_size}
\end{table}


\subsubsection{Effects of the Recurrent Shape Features}
In our second set of experiments, we evaluate the advantages of the proposed recurrent shape features in our framework. To make a fair comparison, we compare two methods:
\begin{itemize}
\item \textbf{facial landmark detection with recurrent shape features}. We use the recurrent shape features in our network as shown in Fig.  \ref{cnn_rnn} (B).

\item \textbf{facial landmark detection without recurrent shape features}. The facial landmark detection is learned without the assistance of the recurrent shape features, i.e., only use the shape-indexed features as shown in Fig. \ref{cnn_rnn} (A).
\end{itemize}

Since the two methods use the same network and the only different is that using or not using the recurrent shape features, these comparisons can show us whether the recurrent shape features can contribute to the accuracy or not.

\begin{table}[h]
\small
    \centering \caption{Mean Errors on 300-W dataset.}
    \begin{tabular}{c|c c c}
      \hline
    & Common & Challenging & Fullset \\
        \hline
          \multicolumn{4}{c}{VGG-S (128x128)}\\
          \hline
         without recurrent features  & 5.19 & 9.75 & 6.08 \\
         with recurrent features  & \textbf{4.62} & \textbf{9.01} & \textbf{5.48} \\
         \hline
          \multicolumn{4}{c}{VGG-S (256x256)}\\
          \hline
       without recurrent features  & 4.51 & 9.64 & 5.51 \\
       with recurrent features  &\textbf{ 4.22} & \textbf{8.23} &\textbf{5.01} \\
        \hline
         \multicolumn{4}{c}{VGG-19 (128x128)}\\
          \hline
        without recurrent features & 4.81 & 9.22 & 5.67 \\
        with recurrent features &\textbf{ 4.57} &\textbf{8.80} & \textbf{5.40} \\
        \hline
          \multicolumn{4}{c}{VGG-19 (256x256)}\\
          \hline
        without recurrent features & 4.19 & 8.42 & 5.02\\
        with recurrent features &\textbf{ 4.07} & \textbf{8.29} & \textbf{4.90}\\
        \hline
        \end{tabular}
    \label{compare_rsf}
\end{table}

Table \ref{compare_rsf} shows the comparison results with respect to Mean Errors. The results show that our proposed recurrent shape features can achieve better performance than the baseline that are without recurrent shape features, especially for the small network (VGG-S). For instance, the mean errors of fullset is 5.48 when the sub-network is VGG-S and the image size is $128 \times 128$, compared to 6.08 of the baseline. The main reason is that the proposed features can learn the temporal dynamics information.

\subsubsection{Effects of the Local Mapping Functions} \label{Eff_LMF}

\begin{table}[h]
\small
    \centering \caption{Comparison with mean shape initialisation.}
    \begin{tabular}{c|c c c}
      \hline
 & \multicolumn{3}{|c}{300-W}\\
    & Common & Challenging & Fullset \\
        \hline
         Mean Shape  & 4.31 & 8.29 & 5.08  \\
         \hline
        N = 50 & 4.07 & 8.29 & 4.90 \\
        \hline
        N = 500 & 4.06  & 8.29 & 4.89 \\
        \hline
        N = 5000 & 4.07 & 8.32 & 4.91 \\
        \hline
        \end{tabular}
    \label{mean_shape}
\end{table}

In this set of experiments, we show the advantages of the proposed local mapping functions. To give an intuitive comparison, we compare with the mean shape initialization setting as shown in Fig. \ref{framework}. We use the same network and same cascade regressions. The only different is that we use the mean shape as the initialisation instead of using the local mapping functions to search the initialisation. These comparisons can answer us whether the proposed local mapping functions can contribute to improve the accuracy or not. We also added experiments to explore the effect of different number of candidate shapes $N$.

Table \ref{mean_shape} show the comparison results. The results show that the proposed local mapping functions perform better that the mean shape. There are two main reasons for the good performance. First, we learn local mapping functions together with the cascaded regression, which means two correlated tasks are learned together. Zhang et al. \cite{zhang2014facial} had showed that leaning related tasks can improve the detection robustness. The second reason is that our method can find the better initialisation. 

The results in Table \ref{mean_shape} also show that the method using $N=50$ candidate shapes performs close to those using $N=500$ or $N=5000$ shapes. Thus, we simply use $N=50$ candidate shapes in this paper.


Note that different with the CFSS method which the good initialisation performs significantly better than the mean shape initialisation, our method is not sensitize to the initialisation and performs more stable. The reason maybe that the proposed recurrent shape features can keep the previous information and help to be more robust to the initialization.

\begin{figure}[t]
\centering
    \includegraphics[width=1\hsize \hspace{0.01\hsize}]{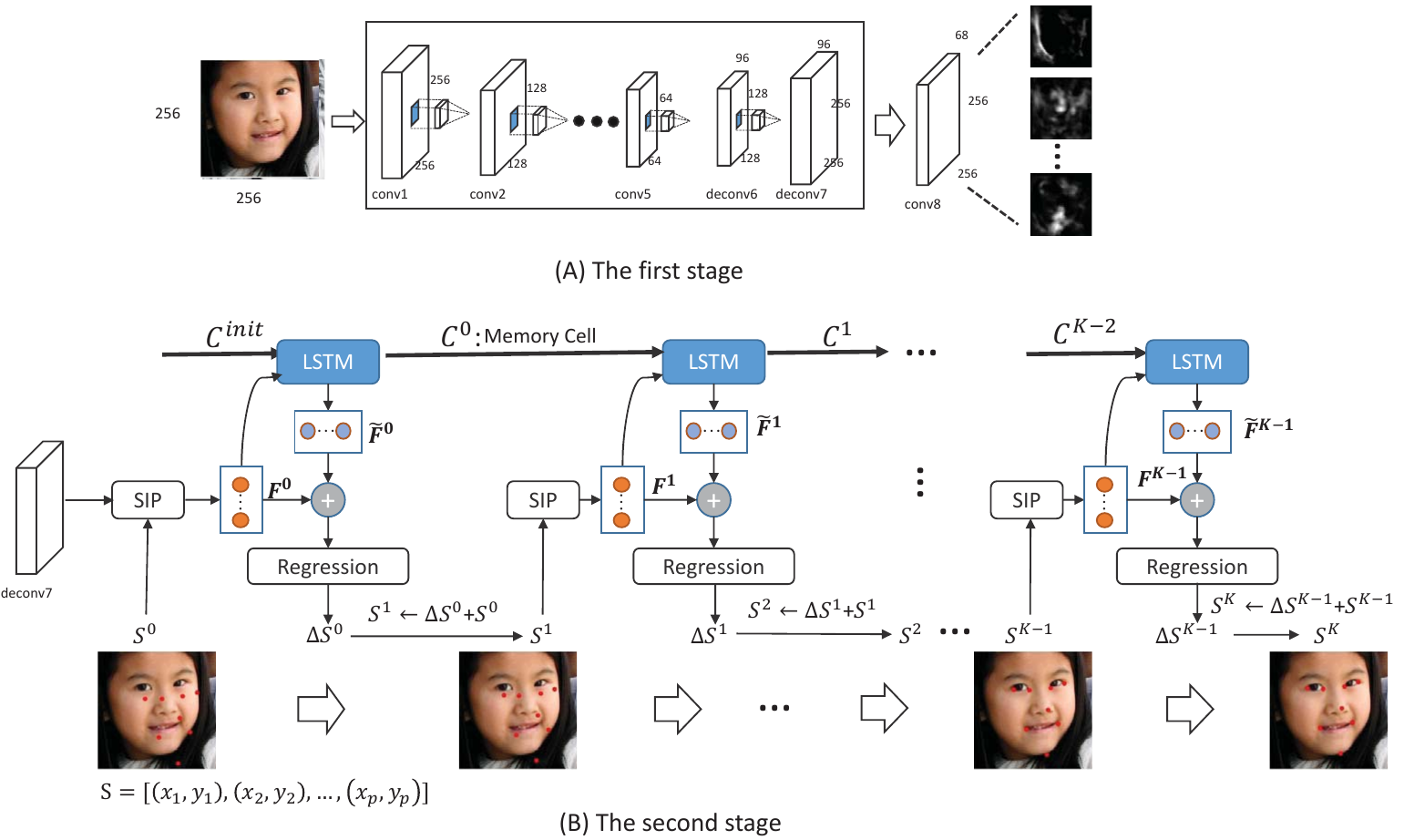}
  \caption{The two-stage learning strategy. }
  \label{two-stages}  
\end{figure}

\subsubsection{Effect of the End-to-end Learning}
Our framework is an end-to-end framework. To show the advantages of the end-to-end framework, we compare to the following baselines. The first baseline is SDM~\cite{sdm}. SDM is a traditional cascaded regression method, which is not an end-to-end framework. The second baseline also uses our framework but adopts a two-stage strategy. We divide our framework into two parts according to the pipeline of the traditional cascaded regression methods: (1) visual features are first extracted; (2) the estimated shapes are then updated via regression from the extracted features; In the first stage, we learn the visual feature as shown in the Fig. \ref{two-stages} (a). In the second stage, we learn the regression via the visual features as shown in Fig. \ref{two-stages} (b). Note that the deconv7 is fixed in the second stage.

\begin{table}[h]
\small
    \centering \caption{Comparison with the non end-to-end methods.}
    \begin{tabular}{c|c c c}
      \hline
 & \multicolumn{3}{|c}{300-W}\\
    & Common & Challenging & Fullset \\
        \hline
         SDM  & 5.57 & 15.40 & 7.50  \\
         \hline
        Two-stage & 4.64 & 8.91 & 5.47 \\
        \hline
        OURS & \textbf{4.07} & \textbf{8.29} & \textbf{4.90} \\
        \hline
        \end{tabular}
    \label{two-stage}
\end{table}

Table \ref{two-stage} shows the comparison results. We can observe that the two-stage method performs better that the SDM, and our one-stage method performs better than the two-stage method. It is desirable to learn the whole facial landmark detection process in the end-to-end framework.

\subsubsection{Effects of the Number of Regressions}
In this paper, we use $K=8$ regressions. Table \ref{steps} shows the mean errors of all these regressions. We can see that the results are very close to each other after 4 steps. $K = 8$ used in this paper is larger enough to let our method converges.
\begin{table}[h]
\small
    \centering \caption{Effects of the number of regressions.}
    \begin{tabular}{c|c c c}
      \hline
 & \multicolumn{3}{|c}{300-W}\\
    & Common & Challenging & Fullset \\
        \hline
         Step 1 &  4.77 & 9.72 & 5.74 \\
         \hline
        Step 2 &  4.21 & 8.69 & 5.08 \\
           \hline
         Step 3 & 4.10 & 8.40 & 4.94  \\
         \hline
        Step 4 &  4.08 & 8.33 & 4.91 \\
           \hline
         Step 5 &  4.08 & 8.31 & 4.91  \\
         \hline
        Step 6 &  4.07 & 8.30 & 4.90 \\
           \hline
         Step 7 &  4.07 & 8.29 & 4.90  \\
         \hline
        Step 8 &  4.07 & 8.29 & 4.90 \\
           \hline
        \end{tabular}
    \label{steps}
\end{table}

\subsubsection{Comparison Results of $5$-point Facial Landmark Detection}
The proposed method can be use predict different number of facial landmarks, e.g., 68 landmark points or 5 landmark points. In order to conduct comparisons with more deep-networks-based methods, we evaluate the performance of the proposed method on datasets of $5$-point facial landmark detection.

We conduct experiments on the Multi-Attribute Facial Landmark (MAFL) dataset. MAFL contains 20,000 facial images randomly chosen from the Celebrity face dataset~\cite{sun2014deep}, each image has a $5$-point landmark annotation. Following the setting of~\cite{zhang2015learning}, $1,000$ faces are used as the test set, and the rest images are used as the training set.

\begin{table}[h]
\small
    \centering \caption{Comparison results w.r.t. mean Error on the MAFL dataset.}
    \begin{tabular}{c c }
        \hline
 \multicolumn{2}{c}{MAFL} \\
Methods & 5 pts \\
        \hline

 CFAN & 15.84 \\

 Cascaded CNN & 9.73   \\

 TCDCN & 7.95   \\
 \hline

 OURS & \textbf{4.51} \\
 \hline
        \end{tabular}
    \label{MAFL}
\end{table}

Three deep-networks-based methods are used as the baselines: CFAN~\cite{zhang2014coarse}, Cascaded CNN~\cite{sun2013deep} and TCDCN~\cite{zhang2015learning}. Table \ref{MAFL} shows the comparison results w.r.t. the mean error on the MAFL dataset. We can see that, in the $5$-point facial landmark detection, the proposed method performs better than the deep-networks-based baselines.

\subsubsection{Effect of Incorporating Shallower-Layer Features}
In the proposed method, we use deconv7 layer as input to the SIP layer to extract deep shape-indexed features. A natural question arising here is whether we can combine deconv7 with the shallower-layer features. To answer this question, we evaluate the performance of different variants of the proposed method. In each variant, we use conv5, deconv6, deconv7, conv5 combined with deconv7, deconv6 combined with deconv7 as the input to the SIP layer for extracting deep shape-indexed features, respectively. The comparison results w.r.t. mean error on the 300-W dataset are shown in Table \ref{feature_layer}. We can see that combining deconv7 with shallower-layers' features can slightly improve the performance. For example, the mean error of the variant that combines deconv6 and deconv7 is 4.85, compared to the 4.90 of the method that only uses deconv7.

\begin{table}[h]
\small
    \centering \caption{Comparison results w.r.t. mean error on 300-W.}
    \begin{tabular}{c|c c c}
      \hline
 & \multicolumn{3}{|c}{300-W}\\
    & Common & Challenging & Fullset \\
        \hline
        conv5 & 4.21  & 8.30  & 5.01  \\
         \hline
        deconv6 & 4.09  & 8.06 & 4.87 \\
         \hline
        deconv7 & 4.07 & 8.29 & 4.90  \\
         \hline
        conv5 + deconv7 & 4.11  & 8.05 & 4.88 \\
         \hline
        deconv6 + deconv7 & 4.08  & 8.01 & 4.85 \\
         \hline
        \end{tabular}
    \label{feature_layer}
\end{table}


\section{Conclusions}
\label{conclusion}
In this paper, we proposed an end-to-end deep-network-based cascaded regression method for facial landmark detection. In the proposed deep architecture, an input image is firstly encoded into high level descriptors in the same size of the input image. Based on this representation, we proposed to learn a probability map for each facial key point and use these probability maps to find the initialization for the cascaded regression. And then, we proposed two strong features. One is the deep shape-indexed features, which are extracted by the designed shape-indexed pooling layer. Another is the recurrent shape features, which is used to learn the connection between the regressions. Finally, the sequential linear regressions are learned to update the shapes. Empirical evaluations on three datasets show that the proposed method significantly outperforms the state-of-the-arts.


%

\appendices



\ifCLASSOPTIONcaptionsoff
  \newpage
\fi



\bibliographystyle{IEEEtran}
\bibliography{IEEEabrv,alignment_bibfile}
\end{document}